\documentclass[10pt,twocolumn,letterpaper]{article}
\usepackage{wacv}
\usepackage[accsupp]{axessibility}
\usepackage{times}
\usepackage{epsfig}
\usepackage{graphicx}
\usepackage{amsmath}
\usepackage{amssymb}
\usepackage{booktabs}

\usepackage{amsmath}
\usepackage{amssymb}

\def\wacvPaperID{754} 

\wacvalgorithmstrack   

\wacvfinalcopy 

\ifwacvfinal
\usepackage[breaklinks=true,bookmarks=false]{hyperref}
\else
\usepackage[pagebackref=true,breaklinks=true,colorlinks,bookmarks=false]{hyperref}
\fi

\begin{document}

\title {Semantics-Depth-Symbiosis: Deeply Coupled Semi-Supervised Learning of
Semantics and Depth}
\author{Nitin Bansal$^1$, Pan Ji\thanks{Worked when associated with OPPO US Research Center} $^{ 1}$, Junsong Yuan$^2$, Yi Xu$^1$\\
$^1$OPPO US Research Center, USA\\
$^2$State University of New York at Buffalo, USA\\
{\tt\small \{nitin.bansal,yi.xu\}@innopeaktech.com, peterji1990@gmail.com, jsyuan@buffalo.edu}
}

\pagestyle{empty}
\maketitle
\thispagestyle{empty}

\begin{abstract}
   Multi-task learning (MTL) paradigm focuses on jointly learning
two or more tasks, aiming for an improvement w.r.t model’s
generalizability, performance, and training/inference memory footprint.
The aforementioned benefits become ever so indispensable in the case of
training for vision-related dense prediction tasks. In this work, we
tackle the MTL problem of two dense tasks, i.e., semantic segmentation
and depth estimation, and present a novel attention module called Cross-Channel Attention Module (CCAM), which facilitates effective feature
sharing along each channel between the two tasks, leading to mutual
performance gain with a negligible increase in trainable parameters. In a symbiotic spirit, we also formulate novel data augmentations for the
semantic segmentation task using predicted depth called AffineMix, and
one using predicted semantics called ColorAug, for depth estimation task.
Finally, we validate the performance gain of the proposed method on the
Cityscapes and ScanNet dataset. which helps us achieve state-of-the-art results for a
semi-supervised joint model based on depth estimation and semantic segmentation.
\end{abstract}

\section{Introduction}
Convolutional Neural Networks (CNNs)~\cite{726791} have helped achieve state-of-the-art results for a range of computer vision tasks including image classification~\cite{he2015deep}, semantic segmentation~\cite{jonathanfcn, chen2017deeplab, chen2017rethinking, li2018pyramid}, and depth estimation~\cite{godard2019digging}. Generally, each of these tasks is trained in isolation, assuming that inter-task features are largely independent. On the contrary, multiple works in the domain of multi-task learning (MTL)~\cite{sun2020adashare, liu2019endtoend,zhang2021survey, Chen_2019_CVPR,Jiao_2018_ECCV, klingner2020selfsupervised,ramirez2018geometry,hoyer2021ways} point towards an inherent symbiotic relation among multiple tasks, where one task benefits from other \textit{sibling} tasks. In~\cite{standley2020tasks}, Standley~\etal{} particularly point towards a synergy between semantic segmentation and depth estimation task, when trained jointly.

Despite previous success, the lack of sufficient labeled data for multi-task training can affect the performance of MTL. To address this limitation, MTL models such as ~\cite{heuer2021multitask,kokkinos2017ubernet,long2017learning, gao2019nddr} propose parameter sharing to overcome data sparsity and enforce generalization by leveraging task losses to regularize each other. The data sparsity problem is more emphatically seen for dense tasks such as semantic segmentation and depth estimation, where obtaining perfect per-pixel annotations is both expensive and untenable in some scenarios, thus fully supervised learning may not be feasible. Motivated by the above observations, we propose to leverage large-scale video data using a semi-supervised MTL training paradigm~\cite{laine2017temporal, tarvainen2018mean, miyato2018virtual,xie2020unsupervised} where semantic segmentation follows a semi-supervised setting and the depth sub-model is trained in self-supervised manner.
Specifically, to improve semi-supervised semantic training, we propose an AffineMix data augmentation strategy, which aims to create new labeled images under a varied range of depth scales. Under this scheme, randomly selected movable objects are projected over the same image, for a randomly selected depth scale. This unlocks another degree of freedom in data augmentation scheme, generating images which are not only close to original data distribution but also more diverse and class balanced. On the depth augmentation end, we propose a simple yet effective data augmentation scheme called ColorAug, which establishes strong contrast between movable objects and adjacent regions, using intermediate semantics information. At last, we employ orthogonal regularization as an strategy to improve  MTL training efficacy, which has a positive impact on both semantics and depth evaluation.

Besides training paradigm, the architecture of MTL model is equally important as it determines how exactly the intermediate features, belonging to different tasks, interact with each other. As highlighted in~\cite{zhang2021survey}, \textit{what to share and how to share} still remains an open question. To improve MTL performance, we adopt a hybrid parameter sharing approach ~\cite{xu2018padnet, hoyer2021ways}, which enforce a soft parameter sharing at decoder layers to facilitate both flexibility and inter-feature learnability. To emphasize inter-channel interactions between tasks, our proposed Cross-Channel Attention Module (CCAM) enforces dual attention on intermediate depth and segmentation features over both spatial and channel dimensions.  This enables us to estimate a degree-of-affinity between inter task channel-features as an intermediary score in an end-to-end differentiable framework. Using CCAM, we can linearly weight the contribution of features from each task before sharing and thus encourage a more informed and reliable feature transfer between two tasks.
\begin{figure*}[!t]
	\begin{center}
		\includegraphics[scale=0.290]{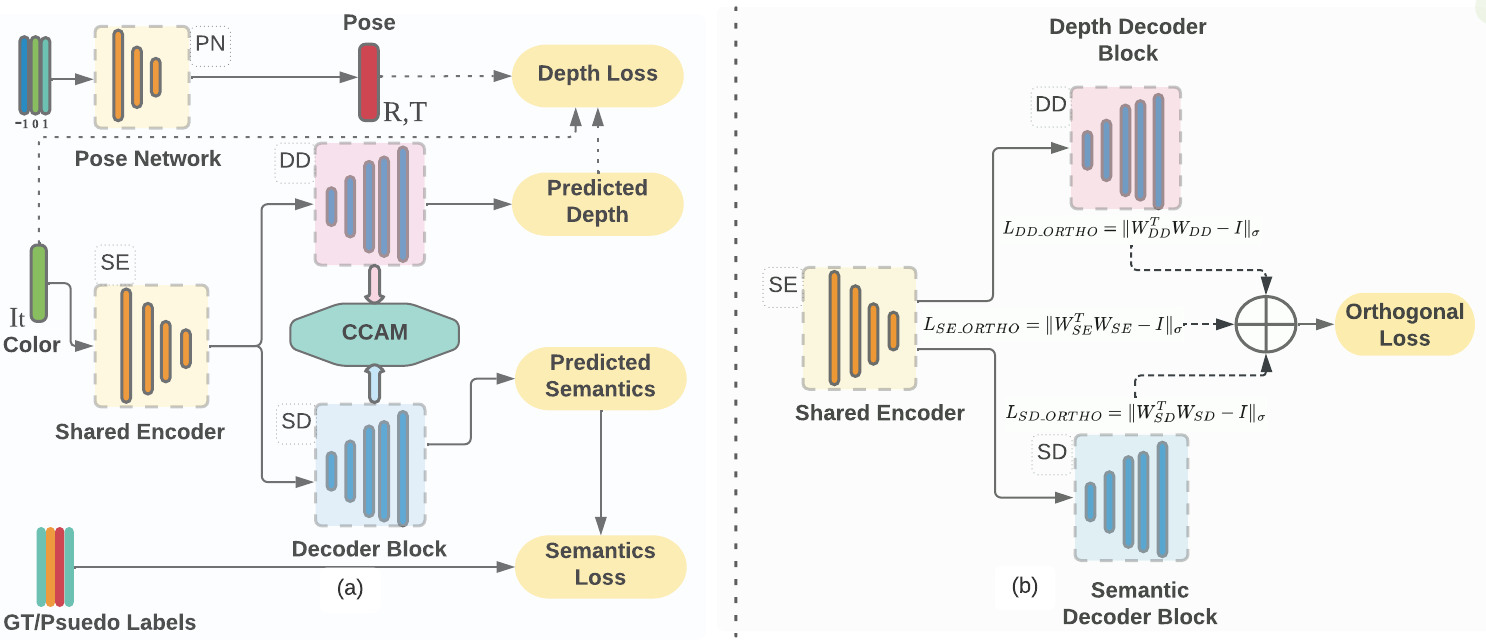}
	\end{center}
    \vspace{-1.0em}
	\caption{(a): Block diagram of the complete multi-task model with semantics and depth losses. (b): Orthogonal regularization enforced on shared encoder(SE), depth decoder(DD) and semantic decoder(SD).}
        \vspace{-1em}
	\label{fig:block_diagram}
\end{figure*}

To summarize our main contributions:
\begin{itemize}
\setlength\itemsep{0em}
	\item To improve feature sharing for MTL, we propose CCAM to estimate cross channel affinity scores between task features. Which enables better inter-task feature transfer.
	\item To deal with data sparsity for MTL, we design a dual data augmentation mechanism for both semantics and depth. Our method encourages diversity, class balance for semantic segmentation, and better region discrimination for depth estimation.
	\item We incorporate orthogonal regularization for depth and semantics with diminishing weighting to facilitate better feature generalization and independence.
\end{itemize}

\section{Related Work}
\subsection{Multi-Task Learning (MTL)}
MTL~\cite{caruana1997multitask} has been used across various tasks, as it improves generalization by transferring domain information of related tasks as an inductive bias.
Previous works such as~\cite{caruana1997multitask, heuer2021multitask, kokkinos2017ubernet,liu2022planemvs} have advocated for hard parameter sharing, where as others ~\cite{long2017learning, gao2019nddr, misra2016cross} advocate soft sharing of parameters, where specific parameters are connected through a learnable link module. For encoder-decoder architectures, previous works such as~\cite{ruder2019latent, misra2016cross, gao2019nddr} are encoder focused, while others~\cite{zhang2019pattern, xu2018padnet, hoyer2021ways} mainly focus on the decoder modules. Our work follows a hybrid approach. We use a shared encoder but employ soft parameter sharing over the decoder module as shown in Fig~\ref{fig:block_diagram}(a). Our approach gravitates more towards the decoder, but improves from the encoder-focused approaches by enabling an efficient cross-task feature sharing mechanism using CCAM module. Work such as~\cite{liu2019end, rebuffi2017learning} employ Residual adapter module, to dedicate a set of model's parameters to each task in a MTL setup; while we improve model's shared representation capability through effective inter-task feature sharing. From optimization point of view, works such as~\cite{chen2018gradnorm, sener2018multi, guo2018dynamic} enforce task balancing by using gradient based multi-objective optimization. We find little to no effect employing such regularization during our training. Finally 
both adversarial training~\cite{sinha2018gradient, maninis2019attentive} and uncertainty ~\cite{kendall2018multitask} are utilized for MTL. Both of them have entirely different motivations compared to our work, and are not necessarily looking to better feature sharing, for two related tasks.

A few other methods~\cite{Chen_2019_CVPR,ramirez2018geometry} leverage semantic labels to jointly train on depth and semantics but mainly to improve on depth results. Marwin \emph{et al}.~\cite{klingner2020selfsupervised} focus on improving depth results specifically for moving objects in a scene using the intermediary semantics information, whereas Jiao \emph{et al}.~\cite{Jiao_2018_ECCV} tackle long-tail distribution in depth prediction domain using semantics. Some of these models are not truly self-supervised for depth estimation or fail to jointly improve both tasks. Another approach uses knowledge distillation~\cite{guizilini2020semanticallyguided,xu2018padnet} from segmentation to guide depth estimation. Furthermore, some other methods~\cite{xu2018padnet,liu2019endtoend} exploit spatial attention~\cite{mnih2014recurrent,wang2018nonlocal} for more effective cross-feature distillation and task related feature extraction. Most recently, Hoyer \emph{et al.}~\cite{hoyer2021ways} propose a joint training network, which has a feature sharing network between the task as proposed in~\cite{xu2018padnet}. Our experiments show that there is little impact on the result of depth metrics upon inclusion/exclusion of this feature sharing module, suggesting ineffective feature transfer between the two tasks. Also~\cite{hoyer2021ways} focuses only on improving  semantic results using depth estimation procedure but not vice-versa. In contrast, CCAM facilitates proportional sharing of features between depth and semantics across each channel (Fig.~\ref{fig:ccam_block})
. Please refer to Sec.\ref{sec:cross_attention} for more details. 

\subsection{Semi-supervised Semantic Segmentation}
Deep convolutional neural network models~\cite{jonathanfcn, sermanet2014overfeat} have been the go-to network for both supervised and semi-supervised semantic segmentation tasks. Subsequent models have improved by leveraging multi-scale input images~\cite{convolutionfeaturekaiming, chen2016attention, hierarichalyann, lin2016refinenet, lin2016efficient}, which capture finer details using multi-scale features. Furthermore, there have been models using feature pyramid spatial pooling~\cite{liu2015parsenet, zhao2017pyramid} and altrous convolutions~\cite{chen2017deeplab, chen2017rethinking, chen2018encoderdecoder, li2018pyramid, yu2016multiscale} to further assist in better per-pixel feature learning and achieve state-of-the-art results. We choose an architecture similar to U-Net~\cite{ronneberger2015unet}, details about which is in Sec. \ref{sec:exps}.

Semi-supervised semantic segmentation training makes use of an unlabeled set of data. Many approaches take image-level labels~\cite{lee2019ficklenet, li2018tell, wei2018revisiting} and class activation maps~\cite{zhou2015learning,wang2020selfsupervised} as a weak supervision signal, which gives marginal assistance in a dense prediction task such as per-pixel segmentation. Methods based on consistency training~\cite{laine2017temporal, tarvainen2018mean, miyato2018virtual,xie2020unsupervised}  use the idea that label space for unlabeled data should remain broadly unchanged after adding noise or perturbation to the input. Ouali~\textit{et al}.~\cite{ouali2020semisupervised} use the same idea but apply perturbation on encoder features instead. Similarly, CutMix~\cite{yun2019cutmix} vouches for stronger augmentations, where crops from input images and pseudo labels are used to generate additional training data. ClassMix~\cite{olsson2020classmix} takes it a step forward by using pseudo labels to get a mix mask, which is then used in consistency training. Our proposed data augmentation is most similar to DepthMix~\cite{hoyer2021ways}, which uses the idea of ClassMix~\cite{olsson2020classmix} but also maintains geometric consistency. Our approach differs from DepthMix~\cite{hoyer2021ways} on three counts. Firstly, we propose a new data augmentation strategy which generates the pseudo labels for selected foreground classes under different randomly selected depth scales, keeping geometric consistency intact. Secondly, we mix foreground masks over the same image and not across the other images for a given batch. Lastly we only consider movable objects as part of the data augmentation with the idea of better handling the intrinsic data bias due to class imbalance. Please refer to Sec.~\ref{sec:data_aug} for more details.

\subsection{Self-supervised Depth Estimation (SDE)}
Depth estimation in the absence of per-pixel ground truth is a well-studied problem. The self-supervised model relies on minimizing the image reconstruction loss, for an input that could either be in stereo-pairs or in a monocular sequence format. Depth estimation under a stereo setting~\cite{godard2019digging, garg2016unsupervised,godard2017unsupervised} mainly focuses on predicting pixel disparity between the pairs and enforcing a consistency between left and right views. In the monocular sequence scenario, methods such as~\cite{godard2019digging, zhou2017unsupervised, garg2016unsupervised, godard2017unsupervised,ji2021monoindoor,zou2020learning,tiwari2020pseudo,ji2022georefine} minimize the photometric reprojection loss during the training phase using the predicted depth and pose. Our depth module largely follows Godard~\textit{et al}.~\cite{godard2019digging}. In addition to the reprojection loss, we also enforce a per-pixel minimum appearance loss and auto-masking which further improves prediction for occluded and stationary pixels.
\section{Proposed Method}
In this section, we start with the basic architecture in Sec.~\ref{sec:basic_arch}. We then discuss in detail about building blocks of CCAM in Sec.~\ref{sec:cross_attention}. Different strategies used for data augmentation for semi-supervised semantics and self-supervised depth network are subsequently discussed in Sec.~\ref{sec:data_aug}. We then briefly discuss the training strategy for depth and semantics part in Sec.~\ref{sec:depth} and Sec.~\ref{sec:semantics} respectively. We conclude this section by presenting how we can effectively apply orthogonal regularization on semantics and depth modules together to further enhance model's performance in Sec.~\ref{sec:ortho_regularization}.
\begin{figure*}[!t]
	\begin{center}
	\includegraphics[width=2.08\columnwidth,,height=4.5cm]{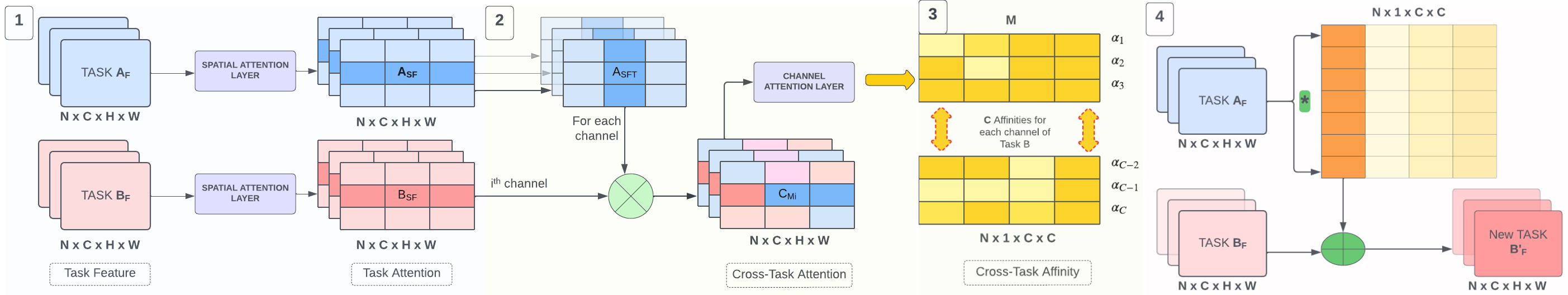}
	\end{center}
    \vspace{-1.0em}
	\caption{Complete sub-blocks of Cross Channel Affinity Module (CCAM). N: Batch Size, C: Number of channels, H: Height, and W: Width.}
        \vspace{-1em}
	\label{fig:ccam_block}
\end{figure*}
\subsection{Basic Architecture}
\label{sec:basic_arch}
For the multi-task training at hand, we follow soft or partial parameter sharing between the tasks. We use a shared encoder but separate decoders for semantics and depth tasks respectively as shown in Fig.~\ref{fig:block_diagram}. Apart from this, we use a separate encoder block for camera pose estimation, and an encoder pretrained on Image{N}et~\cite{ILSVRC15} to calculate the feature distance loss, similar to~\cite{hoyer2021ways}. The CCAM block (details in suppl. material) consists of mainly two sub-blocks, which mainly have convolutional, global average pooling and fully connected layers to compute spatial and cross-channel attention respectively. More specific details about different blocks are mentioned in Sec.~\ref{subsec:exp_details}, under network architecture.
\subsection{Cross Attention Network Architecture}
\label{sec:cross_attention}
Cross-task feature transfer, can be broadly divided in to three sub-categories as described in~\cite{zhang2021survey}: (i) sharing initial layers to facilitate learning common features for complimentary tasks; (ii) using adversarial networks to learn common feature representation as in~\cite{shinohara16b_interspeech}; and (iii) learning different but related feature representations as presented in~\cite{misra2016crossstitch}. 
In this very context, Xu~\etal~\cite{xu2018padnet} propose a multi-modal distillation block, which shares cross-task features through message passing. It simulates a gating mechanism as shown in Eq.~\eqref{eq:feature_share} and ~\eqref{eq:gate}, by leveraging the spatial attention maps of each individual features of all tasks, which then helps decide what features a given task would share with other tasks.
Without loss of generality, suppose we train a total number of $T$ tasks, and ${F_i}^k$ denotes the $i^{th}$ feature of the $k^{th}$ task before message passing and ${F_i}^{o,k}$ after message passing. Xu \textit{et al}.~\cite{xu2018padnet} define the message transfer as:
\begin{equation}
\vspace{-0.5em}
\begin{aligned}
{F_i}^{o,k} = {F_i}^k + \sum_{t=1(\neq k)}^{T} {G_i}^k \odot (W_{t,k} \otimes {F_i}^t)\;,
\end{aligned}
\label{eq:feature_share}
\end{equation}
where $ \odot$ means element-wise product, $\otimes$ represents convolution operation, $W_{t,k}$ represents the convolution block and  ${G_i}^k$ denotes the gating matrix for $i^{th}$ feature of the  $k^{th}$ task:
\begin{equation}
\begin{aligned}
{G_i}^k = \sigma ({W_g}^k \otimes {F_i}^k)\;,
\end{aligned}
\label{eq:gate}
\end{equation}
where ${W_g}^k$ is a convolution block and $\sigma$ denotes the sigmoid operator. We refer reader to~\cite{xu2018padnet} for more details regarding above message passing strategy. According to Eq.~\eqref{eq:feature_share}, it only shares cross-task features naively across the channel dimension. Suppose we are training simultaneously for two tasks namely: \text{$F^k$} and \text{$F^l$}. Eq.~\eqref{eq:feature_share} indirectly implies that $i^{th}$ channel-feature of \textit{$F^k$} is only important to $i^{th}$ channel-feature of \textit{$F^l$}, which is not necessarily true in all scenarios. We overcome this major limitation by designing a module that calculates an affinity vector $\alpha_i$, which gives an estimate about how the $i^{th}$ channel of task \text{$F^k$} is related to any $j^{th}$ channel of task \text{$F^l$}.
As shown in Fig.~\ref{fig:ccam_block}, the entire process of building scores of inter-task channels can be subdivided into four sub-blocks. Since in our case we are dealing with two tasks, we denote them as Task {A} and Task {B}. We extract intermediate output features from respective decoder modules represented by {$A_F$} and {$B_F$} respectively, where $A_F$, $B_F$ $\in$ $\mathbb{R}^ {N \times C \times H \times W}$.\\
\noindent {\bf CCAM Sub-block 1:} We start with sub-block 1, where task's intermediate features \textit{$A_F$} and \textit{$B_F$} are passed through a sequence of conv-blocks ($W_A$ and $W_B$), which serves as spatial attention layers, to compute  \textit{$A_{SF}$} and \textit{$B_{SF}$} according to the following equations:
\begin{equation}
\begin{aligned}
{A_{{SF}_{i}}} = \sigma ({W_A} \otimes {A_{{F}_i}})\;,
\end{aligned}
\label{eq:ASF}
\end{equation}
\begin{equation}
\begin{aligned}
{B_{{SF}_{i}}} = \sigma ({W_B} \otimes {B_{{F}_i}})\;.
\end{aligned}
\label{eq:BSF}
\end{equation}
The idea is to get much more refined features from both tasks before estimating their cross-correlation. The output of this layer preserves the spatial resolution of the input features and gives output features represented by $A_{SF}$ and $B_{SF}$ respectively.

\noindent {\bf CCAM Sub-block 2:} Subsequently in sub-block 2, we build a cross-task relation matrix $C_{Mi}$ for each channel \textit{i} of \textit{$A_{SF}$}, where $C_{Mi}\in \mathbb{R}^{B \times C \times H \times W}$. We then pass the resultant matrix $C_{Mi}$ to a channel attention module $\Psi$, which estimates the affinity vector $\alpha_{i}$ between $i^{th}$ channel of  \textit{ $A_{SF}$} and all the channels of \textit{$B_{SF}$} in accordance with the equation:
\begin{equation}
\begin{aligned}
\alpha_i = \Psi ({A_{{SF}_{i}}} \otimes (B_{SF})^T)\;,
\end{aligned}
\label{eq:CHNATTN}
\end{equation}
where $\Psi$ denotes a combination of global average pooling layer followed by fully connected layers, with a sigmoid layer at the end, which serves as channel attention layer. We repeat this for all the channels of ${A_{SF}}$ to get the corresponding affinity vector $\alpha$.\\
\noindent {\bf CCAM Sub-block 3:} As part of sub-block 3, we accumulate \textit{Affinity Scores} for all the channels of \textit{$A_F$} to achieve a final cross task affinity matrix \textit{$M$}, given by the equation:
\begin{equation}
\begin{aligned}
M = [\alpha_i \oplus \alpha_j]  \quad \forall i, j \in [0, C)\;,
\end{aligned}
\label{eq:ATTN}
\end{equation}
where $\oplus$ denotes concatenation across the \text{row} dimension. \\
\noindent {\bf CCAM Sub-block 4:} Finally in sub-block 4, the cross task affinity matrix \textit{$M$} achieved serves as a score accumulator, which helps get a linearly weighted features ${A'}_F$ and ${B'}_F$, given by equation:
\begin{equation}
\begin{aligned}
{A'}_F = A_F + (B_F \odot M)\;,
{B'}_F = B_F + (A_F \odot M^T)\;,
\end{aligned}
\label{eq:ATTN}
\end{equation}
where $\odot$ represents element-wise multiplication.
\subsection{Data Augmentation}
\label{sec:data_aug}
Data augmentation plays a pivotal role in all machine learning tasks, as it helps gather varied data samples from a similar distribution. In the spirit of cooperative multi-tasking, in this subsection, we introduce novel ways of data augmentation on both segmentation and depth estimation tasks using predicted depth and semantics respectively.

\begin{figure*}[!t]
	\includegraphics[scale=0.284]{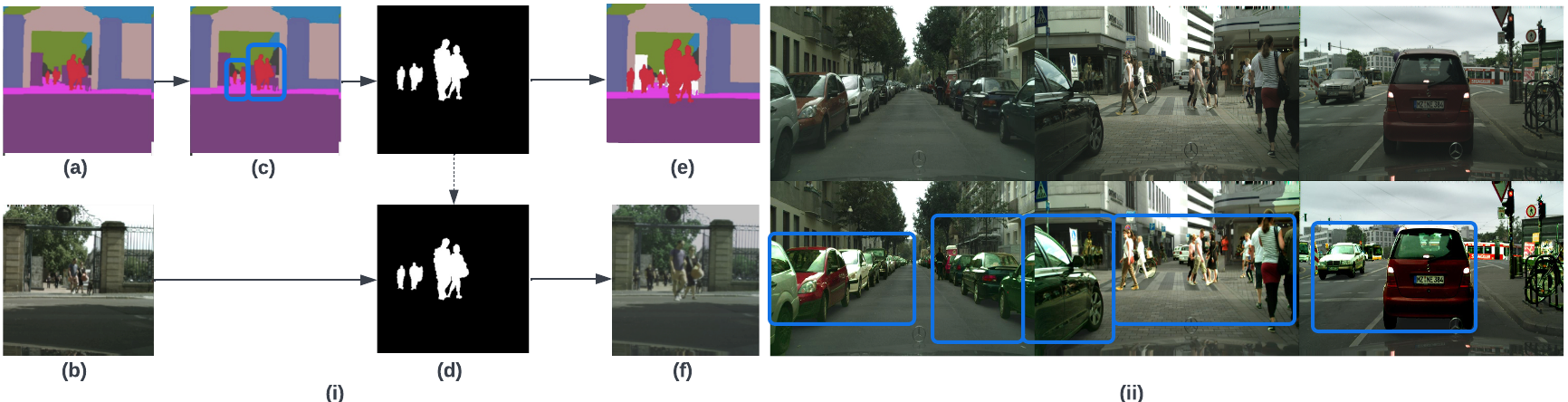}
    \vspace{-0.3em}
	\caption{(i) \textbf{AffineMix for Semantic Segmentation:} Steps involved : (a) GT/ Generated Psuedo label image; (b) Input image; (c) Randomly selected movable object (\textbf{`Person'} class) with random depth scaling of \textbf{0.75}; (d) Occlusion aware, affine generated foreground mask; (e) Augmented label image; (f) Augmented Image. (ii) \textbf{ColorAug for Depth:} Using intermediate semantics output, purposefully create regions of different brightness, contrast and saturation around movable objects highlighted using blue boxes. Row 1: Original Image, Row 2: Augmented Image.}
	\vspace{-3.2mm}
	\label{fig:data_aug}
\end{figure*}

\noindent {\bf Data Augmentation for Segmentation}
 In the context of semi-supervised semantic segmentation, models such as \cite{olsson2020classmix,yun2019cutmix, french2020semisupervised} leverage consistency training by mixing image masks across two different images to generate a new image and its semantic labels. Hoyer \emph{et al.}~\cite{hoyer2021ways} go a step further to generate a much more diverse mixed label space by maintaining the integrity of the scene structure. We propose a new version of data augmentation called AffineMix, which considers mixing labels within the same image under a varied range of random depth values, thus producing a new set of affine-transformed images (see Fig~\ref{fig:data_aug}(i)).
To further improve the data augmentation process, we mix masks associated with only movable objects, to counter class imbalance, which is stark in a dataset such as Cityscapes.
 Given an image \textit{I} and corresponding predicted depth \textit{D}, we seek to generate a mixed image \textit{I'}, by scaling the depth of a selected movable object \textit{m} by a scale factor of \textit{s}:
\begin{equation}
\begin{aligned}
D' = s * D\;,
\end{aligned}
\label{depth_scaling}
\end{equation}
such that its spatial location in the image is changed in a geometrically realistic way.
Changing the depth by a factor of \textit{s}, results in an inverse scaling in the image domain and translational shift which would be given by:
\begin{equation}
\begin{aligned}
t_x = (1.0 - 1/s) * o_x\;,
\end{aligned}
\label{trans_x}
\end{equation}
\begin{equation}
\label{trans_y}
\begin{aligned}
t_y = (1.0 - 1/s) * o_y\;,
\end{aligned}
\end{equation}
where $o_x$ and $o_y$ are normalized offsets along x and y directions. Using $t_x$, $t_y$ and inverse scaling $1/s$, we can perform affine transformation on the image and label space to generate \textit{$I_a$} and \textit{$L_a$}. We then estimate the foreground mask, by comparing the new and old depths and masking it with the region which has the movable object in \textit{$I_a$} and name it \textit{$M_m$}. The final image and label would be then given by:
\begin{equation}
\begin{aligned}
\textit{I'} = \textit{$M_m$} \odot \textit{$I_a$} + (1-\textit{$M_m$}) \odot \textit{$I$}\;,
\end{aligned}
\label{Image_final}
\end{equation}
\begin{equation}
\begin{aligned}
\textit{L'} = \textit{$M_m$} \odot \textit{$L_a$} + (1-\textit{$M_m$}) \odot \textit{$L$}\;.
\end{aligned}
\label{Label_final}
\end{equation}
\noindent{\bf Data Augmentation for Depth}
 As shown in work~\cite{jjgibson, 89ea88f7802a44c89dd104091cc59f75}, factors such as position in the image, texture density, shading, and illumination are some of the pictorial cues about distance in a given image. Recent work in this field~\cite{vandijk2019neural} also re-emphasizes the importance of contrast between adjacent regions as well as bright and dark regions within an image. We particularly leverage this simple albeit important observation to develop an effective data augmentation technique called ColorAug, which uses different appearance based augmentation on movable and non-movable objects. We use the intermediate semantic labels predicted by the model to guide us in developing such an data augmentation as shown in Fig.~\ref{fig:data_aug}(ii).
\subsection{Self-Supervised Depth}
\label{sec:depth}
Training formulation of the self-supervised depth network mainly follows good practices of~\cite{godard2019digging} in terms of using a per-pixel minimum appearance, reprojection loss, and an auto-masking strategy. Main backbone of the depth network comprises of an encoder-decoder structure, represented by \textit{$SE$} and \textit{$DD$} in Fig.~\ref{fig:block_diagram}. We use a subsidiary pose network \textit{$PN$}, to predict the relative translation (T) and rotation (R) of the source frames \text{$I_{t-1}$} and \text{$I_{t+1}$} with respect to the target frame \text{$I_t$}. Predicted poses and depth are then used to estimate the self-supervised depth loss, denoted by $L_{D}$. Building blocks of the encoder and decoder are specified in more details in Sec.~\ref{subsec:exp_details}. 
\subsection{Semi-Supervised Semantic Segmentation}
\label{sec:semantics}
For the semi-supervised semantic segmentation module, we start with a set of labeled images ${\Omega_L}$, unlabeled images ${\Omega_U}$, and $N$ unlabeled image sequences. We pretrain the pose network $PN$, shared encoder $SE$, and depth decoder $DD$ (see Fig.~\ref{fig:block_diagram}) modules with $N$ unlabeled image sequences, in a similar fashion as mentioned in~\cite{hoyer2021ways}. During the joint-training step, parameters of the depth decoder ($DD$) are used to initialize the segmentation decoder ($SD$) module. The CCAM module springs into action during this stage of training, facilitating informative features transfer between $DD$ and $SD$ modules.
During semi-supervised training, suppose the labeled and unlabeled samples are represented as ($I_L$, $S_L$) and ($I_U$, $S_U$), where $S_U$ represents pseudo labels that are generated using a mean teacher algorithm~\cite{tarvainen2018mean}. The parameters of the mean teacher $\theta_T$ is given by the following equation:
\begin{equation}
\label{eq:teacher}
\begin{aligned}
\theta_T' = \alpha\theta_T + (1-\alpha)\theta_{SD}\;,
\end{aligned}
\end{equation}
where $\theta_{SD}$ and $\alpha$ represent parameters of the segmentation decoder module and smoothing coefficient hyper-parameter respectively.
We then generate the pseudo labels $S_U$ as suggested in~\cite{olsson2020classmix} as:
\begin{equation}
\label{eq:unlabeled}
\begin{aligned}
S_U = \underset{l}{\arg\max}(\theta_T(I_U))\;,
\end{aligned}
\end{equation}
where $l$ represents all possible classes in the Cityscapes dataset. Using Eq.~\eqref{eq:teacher} and \eqref{eq:unlabeled}, we can formalize the total semi-supervised loss as:
\begin{equation}
\label{eq:sslloss}
\begin{aligned}
L_{SSL} = L_{CE}(\theta_{SD}(I_L), S_L) + L_{CE}(\theta_{SD}(I_U), S_U)\;.
\end{aligned}
\end{equation}
We then use our AffineMix samples, as a substitute for $I_U$ and $S_U$ in Eq.~\eqref{eq:sslloss} to get the final semi-supervised loss.
\subsection{Orthogonal Regularization (OR)}
Orthogonality constraint on model's parameters has shown encouraging results for tasks such as image classification~\cite{bansal2018gain,Wang_2020_CVPR,8877742}, image retrieval~\cite{Wang_2020_CVPR}, 3D classification~\cite{qi2017pointnet} to name a few.  
Enforcing orthogonality has also helped improve model's convergence, training stability, and promote learning independent parameters. In a multi-task setup such as ours, feature independence within a given task is also important. We study the effect of applying a variation of the orthogonal scheme proposed in~\cite{bansal2018gain} on different sub-modules. The new loss function of the model, after adding orthogonal constraint is given by:
\begin{equation}
\begin{aligned}
L_I = L_{SSL} + L_{D},
\end{aligned}
\label{Label_initial}
\end{equation}
\begin{equation}
\begin{aligned}
L_F = L_I + \lambda\cdot \|W^TW - I\|_\sigma\;,
\end{aligned}
\label{Label_final}
\end{equation}
where $W$, $\|\cdot\|_\sigma$, $I$, $L_F$ and $L_I$ represent the weights (for each layer), spectral norm, identity matrix, final model loss, and initial loss respectively. 
We find that enforcing orthogonality, particularly on the parameters of shared encoder \textit{SE}, depth decoder \textit{DD}, and segmentation decoder \textit{SD} has the most positive impact on the model's performance. We confirm this by calculating average inter-channel correlation, for all decoder layers for both the tasks, with and without OR (more details in supplementary material). 
We postulate that independent features within semantics and depth module would make feature transfer between the tasks more effective. Details about enforcing this regularization and ablation study based on this are provided in Sec.~\ref{subsec:exp_details}.
\label{sec:ortho_regularization}
\begin{figure*}[!t]
		\includegraphics[scale=0.772]{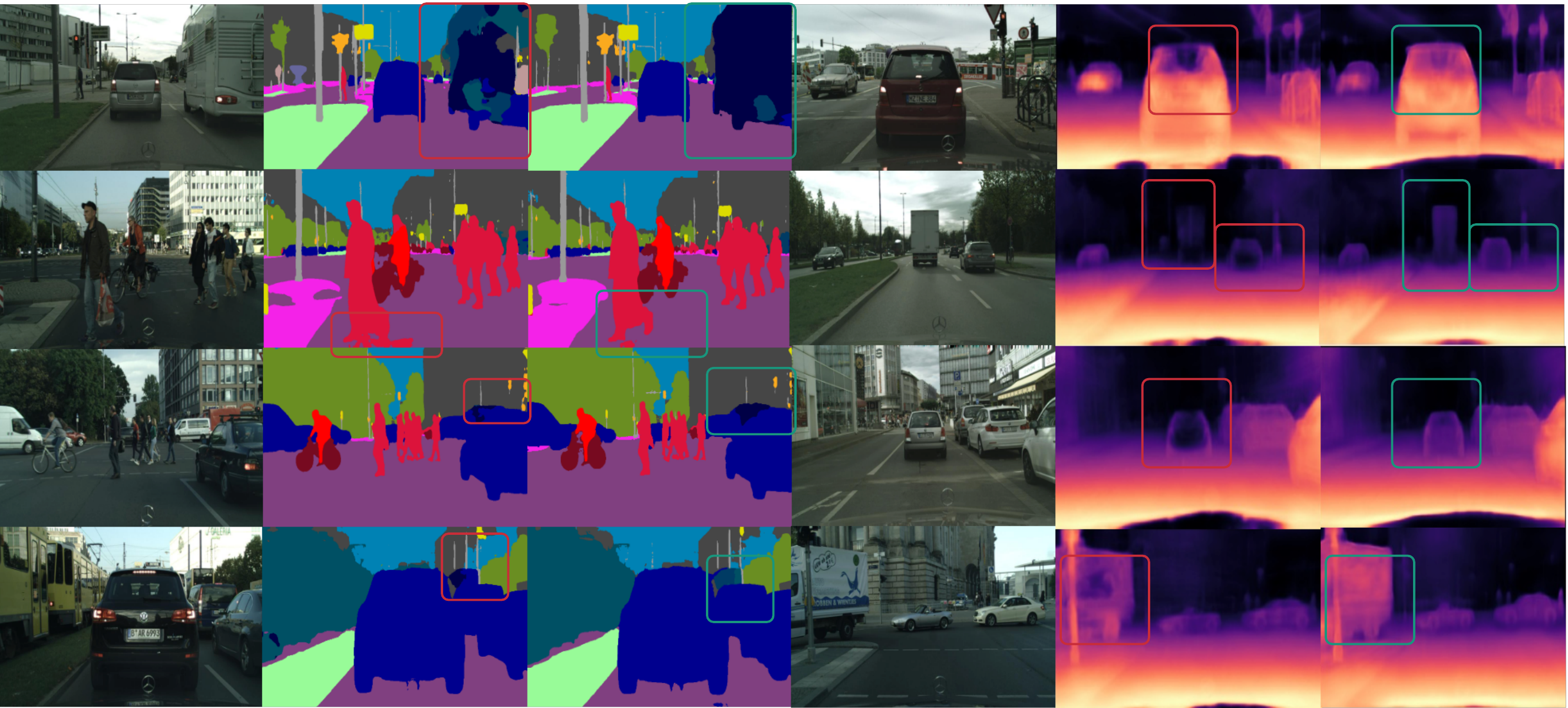}
    \vspace{-0.02em}
	\caption{\textbf{Qualitative results:} In each row we compare the semantics and depth results with the baseline ~\cite{hoyer2021ways}. Red boxes identify shortcomings in the baseline method whereas green boxes highlight the corresponding improvement by using our method.}
	\label{fig:mIoU_comp}
	 \vspace{-0.2em}
\end{figure*}
\section{Experiments}
\label{sec:exps}

\begin{table}[!t]
    \center
    \scalebox{1.0}{
	\begin{tabular}{p{2.4cm} p{1.2cm} p{1.2cm}p{1.2cm}}
		\hline
		Model& 1/8(372) & 1/4(744) & Full(2975) \\
		\hline
		Adversarial\cite{hung2018adversarial} & 58.80& 62.30& \_\\
		s4GAN \cite{mittal2019semisupervised} &59.30&61.90&65.80 \\
		DST-CBC~\cite{feng2021dmt} &60.50&64.40& \_\\
		\hline
		CutMix\cite{french2020semisupervised}   &60.34& 63.87& 67.68\\
		ClassMix \cite{olsson2020classmix} &61.25 &63.30& \_ \\
		\hline
		3-Ways \cite{hoyer2021ways} &68.01 &69.38 &71.16\\
		Ours.    &\textbf{70.72} &\textbf{71.65} & \textbf{72.93}\\
		\hline
	\end{tabular} }
	\vspace{1.5mm}
    \caption{Comparative mIoU (in \%) numbers on Cityscapes Validation set. Reported numbers are mean over three runs, under same experimental settings.}
    \label{table:our_final_table}
    \vspace{-6mm}
\end{table}
\subsection{Experimental Details}
\label{subsec:exp_details}
\noindent{\bf Dataset}
We use the Cityscapes dataset~\cite{cordts2016cityscapes} and ScanNet dataset~\cite{dai2017scannet} for training and evaluating the model. For evaluating on semantics we use the ground-truth labels provided as part of the datasets. We use the data prepossessing and data augmentation step as suggested in~\cite{hoyer2021ways}, where we downsample and center-crop the original 1024x512 color images to 512x512 images for training. For depth we use the unlabeled frames provided by the Cityscapes dataset during the training phase, whereas depth is evaluated against 1,525 images from 6 cities taken from the test set, for which we use the ground-truth depths generated by Watson \etal in~\cite{watson2021temporal}, using the disparity maps and for ScanNet we continue to use the provided ground-truth. More details regarding it is provided in the suppl. material.

\begin{table*}[!t]
\scalebox{0.97}{
\setlength{\tabcolsep}{12.4pt}{
\begin{tabular}{l|c|cccccc}
\hline
Model       &     \multicolumn{1}{c|}{Seg. Metrics} & \multicolumn{6}{c}{Depth Metrics}\\
\hline
& mIoU$\uparrow$ & AbsRel$\downarrow$ &SqRel$\downarrow$& RMSE$\downarrow$& a1$\uparrow$ & a2$\uparrow$ & a3$\uparrow$ \\
\hline 
3-ways* \cite{hoyer2021ways} &68.09&0.150&2.032&7.492&0.824&0.953&0.985\\ 
3-ways*\cite{hoyer2021ways} w/o attn &68.07&0.152&2.497&7.621&\text{0.824}&0.952&0.982\\
Ours (CCAM) &69.35 &\textbf{0.142}&\text{1.653}&\textbf{7.230}&0.824&\textbf{0.957}&\textbf{0.988}\\
Ours + AM &69.84 &0.149&1.651&7.521&0.817&0.952&0.984\\
Ours + AM + OR &\textbf{70.72} &0.146&\textbf{1.546}&\text{7.239}&0.815&\text{0.953}&\text{0.987}\\
Ours + AM + OR + CA &\text{70.20} &\textbf{0.142}&{1.553}&\text{7.284}&\textbf{0.824}&\text{0.956}&\text{0.987}\\
\hline
\end{tabular}}}
\vspace{1mm}
\caption{Comparative mIoU and depth results between baseline model with (vanilla) attention, without attention, and with cross channel affinity attention. Models with * show the results as reproduced by us running the original model. AM: AffineMix, OR: orthogonal Regularization, CA: ColorAug.}
\label{tab:compare_methods}
\end{table*}
\begin{table*}[!t]
    \center
    \scalebox{0.9}{
	\begin{tabular}{p{2.1cm} p{2.1cm} p{2.1cm}p{2.1cm}p{2.1cm} p{2.1cm}}
		\hline
		Metric & CutOut & CutMix & ClassMix & DepthMix &  AffineMix \\
		\hline
		{mIoU} & {57.74} & {60.34} & {63.86} & {68.09} & \textbf{68.70} \\
		\hline
	\end{tabular} }
	\vspace{2mm}
    \caption{Table presents the comparative performance of AffineMix method with previous semi-supervised works, which establishes AffineMix superiority over previous approaches on Cityscapes dataset(1/8 labeled images).}
    \label{tab:city_aug}
    \vspace{-2mm}
\end{table*}
\noindent{\bf Network Architecture}
Basic network as part of our multi-task training is similar to~\cite{hoyer2021ways,xu2018padnet} as seen in Fig.~\ref{fig:block_diagram}(a), which provides an intuitive and effective network for simultaneous training.
It comprises of a shared encoder network which is ResNet101~\cite{he2015deep} and two separate but identical architecture based on Deeplabv3~\cite{chen2017deeplab} with a U-Net~\cite{ronneberger2015unet} decoder.
For pose network we use ResNet18~\cite{he2015deep}, whereas for the shared encoder network we use  ResNet101~\cite{he2015deep} pretrained on the ImageNet dataset~\cite{ILSVRC15}. We refer the readers to the original paper's~\cite{hoyer2021ways} suppl. section for more details. CCAM block consists of mainly two sub-modules, which are, spatial and channel attention layers. Spatial attention comprises of convolutional blocks with kernel size of 3x3. Where as channel attention consist of a global average layer, and two fully connected layers respectively, followed by a sigmoid activation layer. Finer details about CCAM block architecture are provided in suppl. material.\\
\begin{table*}[!t]
\scalebox{1.0}{
\setlength{\tabcolsep}{12.4pt}{
\begin{tabular}{l|c|cccccc}
\hline
Model       &     \multicolumn{1}{c|}{Seg. Metrics} & \multicolumn{6}{c}{Depth Metrics}\\
\hline
& mIoU$\uparrow$ & AbsRel$\downarrow$ &SqRel$\downarrow$& RMSE$\downarrow$& a1$\uparrow$ & a2$\uparrow$ & a3$\uparrow$ \\
\hline 
3-ways* \cite{hoyer2021ways} &39.05&0.184&0.106&7.492&0.688&0.931&0.986\\ 
Ours (CCAM) &\textbf{41.57} &\textbf{0.174}&\textbf{0.095}&\textbf{7.230}&\textbf{0.715}&\textbf{0.939}&\textbf{0.989}\\
\hline
\end{tabular}}}
\vspace{1mm}
\caption{Comparative mIoU and depth results between baseline model with (vanilla) attention, and with cross channel affinity attention for ScanNet dataset}
\label{tab:compare_methods_scannet}
\vspace{-2mm}
\end{table*}
\noindent{\bf Training}
For most part of the training, we follow the strategy as mentioned by Hoyer \textit{et al}.~\cite{hoyer2021ways}. We first train the self-supervised depth and pose module alone using Adam~\cite{kingma2017adam} till 300K iterations, with an initial learning rate of $10^{-4}$, which is reduced by a factor of 10 using the step-scheduled learning rate mechanism. In the second iteration we only look to finetune the shared encoder with a Image{N}et feature distance for another 50K iterations. During joint training, we use SGD with an initial learning rate of $10^{-3}$ and $10^{-2}$ for encoder and decoder respectively, which are reduced by a factor of 10 after 30K iterations. We refer readers to~\cite{hoyer2021ways} for more details. In our observation, we find freezing depth decoder parameters during data augmentation steps of self-supervised semantic learning leads to better model stability without any adverse effect on semantics or depth results. For orthogonal regularization we follow SRIP~\etal~\cite{bansal2018gain} based method, starting with an initial weight $\lambda_O$ = $10^{-4}$, which is then reduced gradually to $10^{-5}$, $10^{-6}$, and $10^{-7}$ after 10K, 20K and 30K iterations respectively.

\subsection{Ablation Studies}
\noindent{\bf Cross Channel Affinity Block}
We study the effectiveness of the self-attention distillation module used in baseline~\cite{hoyer2021ways} for cross-feature transfer between depth and semantics tasks. We follow that up with an experiment, incorporating the CCAM module and validate its positive impact on both tasks. As shown in Tab.~\ref{tab:compare_methods}, we observe minimal improvement to semantics and depth metrics with and without self-attention based distillation module; with mean IoU and absolute relative errors hovering around 68\% and 15\%. Including just the CCAM module, we see an improvement of \textbf{1.28\%} in mIoU and a drop of \textbf{5.3\%} in the absolute relative depth error.

\noindent{\bf AffineMix Augmentation} We further incorporate the proposed AffineMix data augmentation, mainly with an aim to improve on semi-supervised semantic segmentation training. Through our experiments, we find that AffineMix consistently improves upon the baseline by ~\textbf{1.75\%} respectively as shown in Tab.~\ref{tab:compare_methods}. We see a drop in the depth metrics after applying AffineMix. We postulate this drop in depth performance is due to visual incoherence in the local region adjoining to newly added objects, which might be detrimental to the depth task. We also did a comparative ablation experiment verifying the efficacy of proposed data augmentation with the previous data augmentation approaches. As shown in Tab.~\ref{tab:city_aug}, we observe an improvement of about 8.36\%, 4.64\% and, 0.61\% in mIoU numbers when compared to~\cite{yun2019cutmix, olsson2020classmix, hoyer2021ways} respectively. 

\noindent{\bf Orthogonal Regularization}
For verifying the efficacy of OR (Fig.~\ref{fig:block_diagram}(b)), we start out 
by comparing models trained with and without OR, from the perspective of feature independence after the completion of training. As postulated, we find that the features across all 4 layers of depth and semantics decoder module are much more independent after applying such a regularization.(More details in supplementary material).
We further validate this claim by seeing a consistent improvement in semantics and depth metrics. We see an average gain of about \textbf{0.88\%} over a run of three experiments for AffineMix augmentation, and an improvement of about \textbf{2.63\%} compared to the baseline model. We also observe a marginal boost to depth metrics after inclusion of OR by \textbf{2.01\%}(see Tab.~\ref{tab:compare_methods}).

\noindent{\bf ColorAug Augmentation} For the depth augmentation experiment, we leverage the intermediate semantics information to create a new set of images, which help us create regions of different contrast, brightness, and saturation around movable objects. We use this data augmentation scheme only after 10K steps of training, such that semantic outputs from the network are reliable to some extent (mIoU $\geq$ 60\%). We achieve a reduction of about 2.8\% to reach an absolute relative error of \textbf{14.2\%} as seen in Tab.~\ref{tab:compare_methods}, along with an improvement of about \textbf{2.11\%} for semantics.

We also conduct ablation experiments covering all different permutations of AM, CA and OR along with CCAM module(details in suppl. material), mostly giving incremental improvements, proving utility of each individual part. Overall, as seen in Tab.~\ref{table:our_final_table},  we achieve improvements of about \textbf{2.63\%, 2.27\%, 1.77\%} for semantic metrics for 372, 744, and 2,975 samples of Cityscapes dataset respectively, and depth metrics by \textbf{5.3\%} in parallel. Fig.~\ref{fig:mIoU_comp} highlights qualitative improvement seen for both depth and semantic segmentation network. We took a closer look to narrow down the classes, which are most positively impacted by our training strategy.
We find that much of improvement is mainly seen for \textit{low-mIoU-classes} such as motorcycle, wall, rider, and \textit{movable-classes} such as bicycle, train, truck, and bus. \textit{saturated-class} such as building, vegetation, car, sky, and road shows marginal improvements as these classes already have achieved about 90\% mIoU (Qualitative details in supplementary material).

\noindent{\bf ScanNet Dataset}
At last, as part of verifying generalizibility of our proposed CCAM module, we conduct experiments on ScanNet~\cite{dai2017scannet}, which is an indoor dataset, entirely different from Cityscapes dataset in more than one aspects. We observe that with no obvious changes to our training strategy, simply employing CCAM module, our model improves semantics metrics by \textbf{2.52\%} and depth metrics by \textbf{5.7\%} as shown in Tab.~\ref{tab:compare_methods_scannet}. With more than 3\% mIoU improvement for classes such as \textit{door, box, screen,} and \textit{cabinet}. Due to dearth of space, we provide more details about data pre-processing and structuring, train/val split, and class wise mIoU numbers in the suppl. material.
\vspace{-2mm}
\section{Conclusions}
Through this work, we go on to establish, how effective transfer of features between semantics and depth modules, could result in substantial performance gain for both the tasks, in a semi-supervised setting. We follow this up with an intelligent and diverse data augmentation for both depth and semantics. We hope these encouraging results would further push the research community in working towards finding much more efficient and effective ways for multi-task learning. 
{\small
\bibliographystyle{ieee_fullname}
\bibliography{egbib}

\begin{thebibliography}{10}\itemsep=-1pt

\bibitem{bansal2018gain}
Nitin Bansal, Xiaohan Chen, and Zhangyang Wang.
\newblock Can we gain more from orthogonality regularizations in training deep
  cnns?
\newblock {\em arXiv preprint arXiv:1810.09102}, 2018.

\bibitem{caruana1997multitask}
Rich Caruana.
\newblock Multitask learning.
\newblock {\em Machine learning}, 28(1):41--75, 1997.

\bibitem{chen2017deeplab}
Liang-Chieh Chen, George Papandreou, Iasonas Kokkinos, Kevin Murphy, and Alan~L
  Yuille.
\newblock Deeplab: Semantic image segmentation with deep convolutional nets,
  atrous convolution, and fully connected crfs.
\newblock {\em IEEE Transactions on Pattern Analysis and Machine Intelligence},
  40(4):834--848, 2017.

\bibitem{chen2017rethinking}
Liang-Chieh Chen, George Papandreou, Florian Schroff, and Hartwig Adam.
\newblock Rethinking atrous convolution for semantic image segmentation.
\newblock {\em arXiv preprint arXiv:1706.05587}, 2017.

\bibitem{chen2016attention}
Liang-Chieh Chen, Yi Yang, Jiang Wang, Wei Xu, and Alan~L Yuille.
\newblock Attention to scale: Scale-aware semantic image segmentation.
\newblock In {\em Proceedings of the IEEE Conference on Computer Vision and
  Pattern Recognition}, pages 3640--3649, 2016.

\bibitem{chen2018encoderdecoder}
Liang-Chieh Chen, Yukun Zhu, George Papandreou, Florian Schroff, and Hartwig
  Adam.
\newblock Encoder-decoder with atrous separable convolution for semantic image
  segmentation.
\newblock In {\em Proceedings of the European Conference on Computer Vision},
  pages 801--818, 2018.

\bibitem{Chen_2019_CVPR}
Po-Yi Chen, Alexander~H. Liu, Yen-Cheng Liu, and Yu-Chiang~Frank Wang.
\newblock Towards scene understanding: Unsupervised monocular depth estimation
  with semantic-aware representation.
\newblock In {\em Proceedings of the IEEE/CVF Conference on Computer Vision and
  Pattern Recognition}, June 2019.

\bibitem{chen2018gradnorm}
Zhao Chen, Vijay Badrinarayanan, Chen-Yu Lee, and Andrew Rabinovich.
\newblock Gradnorm: Gradient normalization for adaptive loss balancing in deep
  multitask networks.
\newblock In {\em International Conference on Machine Learning}, pages
  794--803. PMLR, 2018.

\bibitem{cordts2016cityscapes}
Marius Cordts, Mohamed Omran, Sebastian Ramos, Timo Rehfeld, Markus Enzweiler,
  Rodrigo Benenson, Uwe Franke, Stefan Roth, and Bernt Schiele.
\newblock The cityscapes dataset for semantic urban scene understanding.
\newblock In {\em Proceedings of the IEEE Conference on Computer Vision and
  Pattern Recognition}, pages 3213--3223, 2016.

\bibitem{dai2017scannet}
Angela Dai, Angel~X. Chang, Manolis Savva, Maciej Halber, Thomas Funkhouser,
  and Matthias Nie{\ss}ner.
\newblock Scannet: Richly-annotated 3d reconstructions of indoor scenes.
\newblock In {\em Proc. Computer Vision and Pattern Recognition (CVPR), IEEE},
  2017.

\bibitem{convolutionfeaturekaiming}
Jifeng Dai, Kaiming He, and Jian Sun.
\newblock Convolutional feature masking for joint object and stuff
  segmentation.
\newblock In {\em Proceedings of the IEEE Conference on Computer Vision and
  Pattern Recognition}, pages 3992--4000, 2015.

\bibitem{vandijk2019neural}
Tom~van Dijk and Guido~de Croon.
\newblock How do neural networks see depth in single images?
\newblock In {\em Proceedings of the IEEE/CVF International Conference on
  Computer Vision}, pages 2183--2191, 2019.

\bibitem{hierarichalyann}
Clement Farabet, Camille Couprie, Laurent Najman, and Yann LeCun.
\newblock Learning hierarchical features for scene labeling.
\newblock {\em IEEE Transactions on Pattern Analysis and Machine Intelligence},
  35(8):1915--1929, 2012.

\bibitem{feng2021dmt}
Zhengyang Feng, Qianyu Zhou, Qiqi Gu, Xin Tan, Guangliang Cheng, Xuequan Lu,
  Jianping Shi, and Lizhuang Ma.
\newblock Dmt: Dynamic mutual training for semi-supervised learning.
\newblock {\em arXiv preprint arXiv:2004.08514}, 2020.

\bibitem{french2020semisupervised}
Geoff French, Samuli Laine, Timo Aila, Michal Mackiewicz, and Graham Finlayson.
\newblock Semi-supervised semantic segmentation needs strong, varied
  perturbations.
\newblock {\em arXiv preprint arXiv:1906.01916}, 2019.

\bibitem{gao2019nddr}
Yuan Gao, Jiayi Ma, Mingbo Zhao, Wei Liu, and Alan~L Yuille.
\newblock Nddr-cnn: Layerwise feature fusing in multi-task cnns by neural
  discriminative dimensionality reduction.
\newblock In {\em Proceedings of the IEEE/CVF Conference on Computer Vision and
  Pattern Recognition}, pages 3205--3214, 2019.

\bibitem{garg2016unsupervised}
Ravi Garg, Vijay~Kumar Bg, Gustavo Carneiro, and Ian Reid.
\newblock Unsupervised cnn for single view depth estimation: Geometry to the
  rescue.
\newblock In {\em European Conference on Computer Vision}, pages 740--756.
  Springer, 2016.

\bibitem{jjgibson}
James~J Gibson.
\newblock The perception of the visual world.
\newblock 1950.

\bibitem{godard2017unsupervised}
Cl{\'e}ment Godard, Oisin Mac~Aodha, and Gabriel~J Brostow.
\newblock Unsupervised monocular depth estimation with left-right consistency.
\newblock In {\em Proceedings of the IEEE Conference on Computer Vision and
  Pattern Recognition}, pages 270--279, 2017.

\bibitem{godard2019digging}
Cl{\'e}ment Godard, Oisin Mac~Aodha, Michael Firman, and Gabriel~J Brostow.
\newblock Digging into self-supervised monocular depth estimation.
\newblock In {\em Proceedings of the IEEE/CVF International Conference on
  Computer Vision}, pages 3828--3838, 2019.

\bibitem{guizilini2020semanticallyguided}
Vitor Guizilini, Rui Hou, Jie Li, Rares Ambrus, and Adrien Gaidon.
\newblock Semantically-guided representation learning for self-supervised
  monocular depth.
\newblock {\em arXiv preprint arXiv:2002.12319}, 2020.

\bibitem{guo2018dynamic}
Michelle Guo, Albert Haque, De-An Huang, Serena Yeung, and Li Fei-Fei.
\newblock Dynamic task prioritization for multitask learning.
\newblock In {\em Proceedings of the European conference on computer vision
  (ECCV)}, pages 270--287, 2018.

\bibitem{he2015deep}
Kaiming He, Xiangyu Zhang, Shaoqing Ren, and Jian Sun.
\newblock Deep residual learning for image recognition.
\newblock In {\em Proceedings of the IEEE Conference on Computer Vision and
  Pattern Recognition}, pages 770--778, 2016.

\bibitem{heuer2021multitask}
Falk Heuer, Sven Mantowsky, Saqib Bukhari, and Georg Schneider.
\newblock Multitask-centernet (mcn): Efficient and diverse multitask learning
  using an anchor free approach.
\newblock In {\em Proceedings of the IEEE/CVF International Conference on
  Computer Vision}, pages 997--1005, 2021.

\bibitem{hoyer2021ways}
Lukas Hoyer, Dengxin Dai, Yuhua Chen, Adrian Koring, Suman Saha, and Luc
  Van~Gool.
\newblock Three ways to improve semantic segmentation with self-supervised
  depth estimation.
\newblock In {\em Proceedings of the IEEE/CVF Conference on Computer Vision and
  Pattern Recognition}, pages 11130--11140, 2021.

\bibitem{hung2018adversarial}
Wei-Chih Hung, Yi-Hsuan Tsai, Yan-Ting Liou, Yen-Yu Lin, and Ming-Hsuan Yang.
\newblock Adversarial learning for semi-supervised semantic segmentation.
\newblock {\em arXiv preprint arXiv:1802.07934}, 2018.

\bibitem{ji2021monoindoor}
Pan Ji, Runze Li, Bir Bhanu, and Yi Xu.
\newblock Monoindoor: Towards good practice of self-supervised monocular depth
  estimation for indoor environments.
\newblock In {\em Proceedings of the IEEE/CVF International Conference on
  Computer Vision}, pages 12787--12796, 2021.

\bibitem{ji2022georefine}
Pan Ji, Qingan Yan, Yuxin Ma, and Yi Xu.
\newblock Georefine: Self-supervised online depth refinement for accurate dense
  mapping.
\newblock {\em arXiv preprint arXiv:2205.01656}, 2022.

\bibitem{Jiao_2018_ECCV}
Jianbo Jiao, Ying Cao, Yibing Song, and Rynson Lau.
\newblock Look deeper into depth: Monocular depth estimation with semantic
  booster and attention-driven loss.
\newblock In {\em Proceedings of the European Conference on Computer Vision},
  September 2018.

\bibitem{kendall2018multitask}
Alex Kendall, Yarin Gal, and Roberto Cipolla.
\newblock Multi-task learning using uncertainty to weigh losses for scene
  geometry and semantics.
\newblock In {\em Proceedings of the IEEE Conference on Computer Vision and
  Pattern Recognition}, pages 7482--7491, 2018.

\bibitem{kingma2017adam}
Diederik~P Kingma and Jimmy Ba.
\newblock Adam: A method for stochastic optimization.
\newblock {\em arXiv preprint arXiv:1412.6980}, 2014.

\bibitem{klingner2020selfsupervised}
Marvin Klingner, Jan-Aike Term{\"o}hlen, Jonas Mikolajczyk, and Tim
  Fingscheidt.
\newblock Self-supervised monocular depth estimation: Solving the dynamic
  object problem by semantic guidance.
\newblock In {\em European Conference on Computer Vision}, pages 582--600.
  Springer, 2020.

\bibitem{kokkinos2017ubernet}
Iasonas Kokkinos.
\newblock Ubernet: Training a universal convolutional neural network for low-,
  mid-, and high-level vision using diverse datasets and limited memory.
\newblock In {\em Proceedings of the IEEE conference on computer vision and
  pattern recognition}, pages 6129--6138, 2017.

\bibitem{laine2017temporal}
Samuli Laine and Timo Aila.
\newblock Temporal ensembling for semi-supervised learning.
\newblock {\em arXiv preprint arXiv:1610.02242}, 2016.

\bibitem{726791}
Yann LeCun, L{\'e}on Bottou, Yoshua Bengio, and Patrick Haffner.
\newblock Gradient-based learning applied to document recognition.
\newblock {\em Proceedings of the IEEE}, 86(11):2278--2324, 1998.

\bibitem{lee2019ficklenet}
Jungbeom Lee, Eunji Kim, Sungmin Lee, Jangho Lee, and Sungroh Yoon.
\newblock Ficklenet: Weakly and semi-supervised semantic image segmentation
  using stochastic inference.
\newblock In {\em Proceedings of the IEEE/CVF Conference on Computer Vision and
  Pattern Recognition}, pages 5267--5276, 2019.

\bibitem{li2018pyramid}
Hanchao Li, Pengfei Xiong, Jie An, and Lingxue Wang.
\newblock Pyramid attention network for semantic segmentation.
\newblock {\em arXiv preprint arXiv:1805.10180}, 2018.

\bibitem{li2018tell}
Kunpeng Li, Ziyan Wu, Kuan-Chuan Peng, Jan Ernst, and Yun Fu.
\newblock Tell me where to look: Guided attention inference network.
\newblock In {\em Proceedings of the IEEE Conference on Computer Vision and
  Pattern Recognition}, pages 9215--9223, 2018.

\bibitem{8877742}
Shuai Li, Kui Jia, Yuxin Wen, Tongliang Liu, and Dacheng Tao.
\newblock Orthogonal deep neural networks.
\newblock {\em IEEE Transactions on Pattern Analysis and Machine Intelligence},
  43(4):1352--1368, 2021.

\bibitem{lin2016refinenet}
Guosheng Lin, Anton Milan, Chunhua Shen, and Ian Reid.
\newblock Refinenet: Multi-path refinement networks for high-resolution
  semantic segmentation.
\newblock In {\em Proceedings of the IEEE Conference on Computer Vision and
  Pattern Recognition}, pages 1925--1934, 2017.

\bibitem{lin2016efficient}
Guosheng Lin, Chunhua Shen, Anton Van Den~Hengel, and Ian Reid.
\newblock Efficient piecewise training of deep structured models for semantic
  segmentation.
\newblock In {\em Proceedings of the IEEE Conference on Computer Vision and
  Pattern Recognition}, pages 3194--3203, 2016.

\bibitem{liu2022planemvs}
Jiachen Liu, Pan Ji, Nitin Bansal, Changjiang Cai, Qingan Yan, Xiaolei Huang,
  and Yi Xu.
\newblock Planemvs: 3d plane reconstruction from multi-view stereo.
\newblock In {\em Proceedings of the IEEE/CVF Conference on Computer Vision and
  Pattern Recognition}, pages 8665--8675, 2022.

\bibitem{liu2019endtoend}
Shikun Liu, Edward Johns, and Andrew~J Davison.
\newblock End-to-end multi-task learning with attention.
\newblock In {\em Proceedings of the IEEE/CVF Conference on Computer Vision and
  Pattern Recognition}, pages 1871--1880, 2019.

\bibitem{liu2019end}
Shikun Liu, Edward Johns, and Andrew~J Davison.
\newblock End-to-end multi-task learning with attention.
\newblock In {\em Proceedings of the IEEE/CVF conference on computer vision and
  pattern recognition}, pages 1871--1880, 2019.

\bibitem{liu2015parsenet}
Wei Liu, Andrew Rabinovich, and Alexander~C Berg.
\newblock Parsenet: Looking wider to see better.
\newblock {\em arXiv preprint arXiv:1506.04579}, 2015.

\bibitem{jonathanfcn}
Jonathan Long, Evan Shelhamer, and Trevor Darrell.
\newblock Fully convolutional networks for semantic segmentation.
\newblock In {\em Proceedings of the IEEE Conference on Computer Vision and
  Pattern Recognition}, pages 3431--3440, 2015.

\bibitem{long2017learning}
Mingsheng Long, Zhangjie Cao, Jianmin Wang, and Philip~S Yu.
\newblock Learning multiple tasks with multilinear relationship networks.
\newblock {\em Advances in neural information processing systems}, 30, 2017.

\bibitem{maninis2019attentive}
Kevis-Kokitsi Maninis, Ilija Radosavovic, and Iasonas Kokkinos.
\newblock Attentive single-tasking of multiple tasks.
\newblock In {\em Proceedings of the IEEE/CVF Conference on Computer Vision and
  Pattern Recognition}, pages 1851--1860, 2019.

\bibitem{misra2016cross}
Ishan Misra, Abhinav Shrivastava, Abhinav Gupta, and Martial Hebert.
\newblock Cross-stitch networks for multi-task learning.
\newblock In {\em Proceedings of the IEEE conference on computer vision and
  pattern recognition}, pages 3994--4003, 2016.

\bibitem{misra2016crossstitch}
Ishan Misra, Abhinav Shrivastava, Abhinav Gupta, and Martial Hebert.
\newblock Cross-stitch networks for multi-task learning.
\newblock In {\em Proceedings of the IEEE Conference on Computer Vision and
  Pattern Recognition}, pages 3994--4003, 2016.

\bibitem{mittal2019semisupervised}
Sudhanshu Mittal, Maxim Tatarchenko, and Thomas Brox.
\newblock Semi-supervised semantic segmentation with high-and low-level
  consistency.
\newblock {\em IEEE Transactions on Pattern Analysis and Machine Intelligence},
  2019.

\bibitem{miyato2018virtual}
Takeru Miyato, Shin-ichi Maeda, Masanori Koyama, and Shin Ishii.
\newblock Virtual adversarial training: a regularization method for supervised
  and semi-supervised learning.
\newblock {\em IEEE Transactions on Pattern Analysis and Machine Intelligence},
  41(8):1979--1993, 2018.

\bibitem{mnih2014recurrent}
Volodymyr Mnih, Nicolas Heess, Alex Graves, et~al.
\newblock Recurrent models of visual attention.
\newblock In {\em Advances in Neural Information Processing Systems}, pages
  2204--2212, 2014.

\bibitem{olsson2020classmix}
Viktor Olsson, Wilhelm Tranheden, Juliano Pinto, and Lennart Svensson.
\newblock Classmix: Segmentation-based data augmentation for semi-supervised
  learning.
\newblock In {\em Proceedings of the IEEE/CVF Winter Conference on Applications
  of Computer Vision}, pages 1369--1378, 2021.

\bibitem{ouali2020semisupervised}
Yassine Ouali, C{\'e}line Hudelot, and Myriam Tami.
\newblock Semi-supervised semantic segmentation with cross-consistency
  training.
\newblock In {\em Proceedings of the IEEE/CVF Conference on Computer Vision and
  Pattern Recognition}, pages 12674--12684, 2020.

\bibitem{qi2017pointnet}
Charles~R Qi, Hao Su, Kaichun Mo, and Leonidas~J Guibas.
\newblock Pointnet: Deep learning on point sets for 3d classification and
  segmentation.
\newblock In {\em Proceedings of the IEEE Conference on Computer Vision and
  Pattern Recognition}, pages 652--660, 2017.

\bibitem{ramirez2018geometry}
Pierluigi~Zama Ramirez, Matteo Poggi, Fabio Tosi, Stefano Mattoccia, and Luigi
  Di~Stefano.
\newblock Geometry meets semantics for semi-supervised monocular depth
  estimation.
\newblock In {\em Asian Conference on Computer Vision}, pages 298--313.
  Springer, 2018.

\bibitem{rebuffi2017learning}
Sylvestre-Alvise Rebuffi, Hakan Bilen, and Andrea Vedaldi.
\newblock Learning multiple visual domains with residual adapters.
\newblock {\em Advances in neural information processing systems}, 30, 2017.

\bibitem{ronneberger2015unet}
Olaf Ronneberger, Philipp Fischer, and Thomas Brox.
\newblock U-net: Convolutional networks for biomedical image segmentation.
\newblock In {\em International Conference on Medical Image Computing and
  Computer-Assisted Intervention}, pages 234--241. Springer, 2015.

\bibitem{ruder2019latent}
Sebastian Ruder, Joachim Bingel, Isabelle Augenstein, and Anders S{\o}gaard.
\newblock Latent multi-task architecture learning.
\newblock In {\em Proceedings of the AAAI Conference on Artificial
  Intelligence}, volume~33, pages 4822--4829, 2019.

\bibitem{ILSVRC15}
Olga Russakovsky, Jia Deng, Hao Su, Jonathan Krause, Sanjeev Satheesh, Sean Ma,
  Zhiheng Huang, Andrej Karpathy, Aditya Khosla, Michael Bernstein,
  Alexander~C. Berg, and Li Fei-Fei.
\newblock {ImageNet Large Scale Visual Recognition Challenge}.
\newblock {\em International Journal of Computer Vision}, 115(3):211--252,
  2015.

\bibitem{sener2018multi}
Ozan Sener and Vladlen Koltun.
\newblock Multi-task learning as multi-objective optimization.
\newblock {\em Advances in neural information processing systems}, 31, 2018.

\bibitem{sermanet2014overfeat}
Pierre Sermanet, David Eigen, Xiang Zhang, Micha{\"e}l Mathieu, Rob Fergus, and
  Yann LeCun.
\newblock Overfeat: Integrated recognition, localization and detection using
  convolutional networks.
\newblock {\em arXiv preprint arXiv:1312.6229}, 2013.

\bibitem{shinohara16b_interspeech}
Yusuke Shinohara.
\newblock {Adversarial Multi-Task Learning of Deep Neural Networks for Robust
  Speech Recognition}.
\newblock In {\em Proc. Interspeech 2016}, pages 2369--2372, 2016.

\bibitem{sinha2018gradient}
Ayan Sinha, Zhao Chen, Vijay Badrinarayanan, and Andrew Rabinovich.
\newblock Gradient adversarial training of neural networks.
\newblock {\em arXiv preprint arXiv:1806.08028}, 2018.

\bibitem{standley2020tasks}
Trevor Standley, Amir Zamir, Dawn Chen, Leonidas Guibas, Jitendra Malik, and
  Silvio Savarese.
\newblock Which tasks should be learned together in multi-task learning?
\newblock In {\em International Conference on Machine Learning}, pages
  9120--9132. PMLR, 2020.

\bibitem{sun2020adashare}
Ximeng Sun, Rameswar Panda, Rogerio Feris, and Kate Saenko.
\newblock Adashare: Learning what to share for efficient deep multi-task
  learning.
\newblock {\em arXiv preprint arXiv:1911.12423}, 2019.

\bibitem{tarvainen2018mean}
Antti Tarvainen and Harri Valpola.
\newblock Mean teachers are better role models: Weight-averaged consistency
  targets improve semi-supervised deep learning results.
\newblock {\em arXiv preprint arXiv:1703.01780}, 2017.

\bibitem{tiwari2020pseudo}
Lokender Tiwari, Pan Ji, Quoc-Huy Tran, Bingbing Zhuang, Saket Anand, and
  Manmohan Chandraker.
\newblock Pseudo rgb-d for self-improving monocular slam and depth prediction.
\newblock In {\em European conference on computer vision}, pages 437--455,
  2020.

\bibitem{Wang_2020_CVPR}
Jiayun Wang, Yubei Chen, Rudrasis Chakraborty, and Stella~X. Yu.
\newblock Orthogonal convolutional neural networks.
\newblock In {\em Proceedings of the IEEE/CVF Conference on Computer Vision and
  Pattern Recognition}, June 2020.

\bibitem{wang2018nonlocal}
Xiaolong Wang, Ross Girshick, Abhinav Gupta, and Kaiming He.
\newblock Non-local neural networks.
\newblock In {\em Proceedings of the IEEE Conference on Computer Vision and
  Pattern Recognition}, pages 7794--7803, 2018.

\bibitem{wang2020selfsupervised}
Yude Wang, Jie Zhang, Meina Kan, Shiguang Shan, and Xilin Chen.
\newblock Self-supervised equivariant attention mechanism for weakly supervised
  semantic segmentation.
\newblock In {\em Proceedings of the IEEE/CVF Conference on Computer Vision and
  Pattern Recognition}, pages 12275--12284, 2020.

\bibitem{watson2021temporal}
Jamie Watson, Oisin Mac~Aodha, Victor Prisacariu, Gabriel Brostow, and Michael
  Firman.
\newblock The temporal opportunist: Self-supervised multi-frame monocular
  depth.
\newblock In {\em Proceedings of the IEEE/CVF Conference on Computer Vision and
  Pattern Recognition}, pages 1164--1174, 2021.

\bibitem{wei2018revisiting}
Yunchao Wei, Huaxin Xiao, Honghui Shi, Zequn Jie, Jiashi Feng, and Thomas~S
  Huang.
\newblock Revisiting dilated convolution: A simple approach for weakly-and
  semi-supervised semantic segmentation.
\newblock In {\em Proceedings of the IEEE Conference on Computer Vision and
  Pattern Recognition}, pages 7268--7277, 2018.

\bibitem{89ea88f7802a44c89dd104091cc59f75}
John~T Wixted and Sharon~L Thompson-Schill.
\newblock {\em Stevens' Handbook of Experimental Psychology and Cognitive
  Neuroscience, Language and Thought}, volume~3.
\newblock John Wiley \& Sons, 2018.

\bibitem{xie2020unsupervised}
Qizhe Xie, Zihang Dai, Eduard Hovy, Minh-Thang Luong, and Quoc~V Le.
\newblock Unsupervised data augmentation for consistency training.
\newblock {\em arXiv preprint arXiv:1904.12848}, 2019.

\bibitem{xu2018padnet}
Dan Xu, Wanli Ouyang, Xiaogang Wang, and Nicu Sebe.
\newblock Pad-net: Multi-tasks guided prediction-and-distillation network for
  simultaneous depth estimation and scene parsing.
\newblock In {\em Proceedings of the IEEE Conference on Computer Vision and
  Pattern Recognition}, pages 675--684, 2018.

\bibitem{yu2016multiscale}
Fisher Yu and Vladlen Koltun.
\newblock Multi-scale context aggregation by dilated convolutions.
\newblock {\em arXiv preprint arXiv:1511.07122}, 2015.

\bibitem{yun2019cutmix}
Sangdoo Yun, Dongyoon Han, Seong~Joon Oh, Sanghyuk Chun, Junsuk Choe, and
  Youngjoon Yoo.
\newblock Cutmix: Regularization strategy to train strong classifiers with
  localizable features.
\newblock In {\em Proceedings of the IEEE/CVF International Conference on
  Computer Vision}, pages 6023--6032, 2019.

\bibitem{zhang2021survey}
Yu Zhang and Qiang Yang.
\newblock A survey on multi-task learning.
\newblock {\em IEEE Transactions on Knowledge and Data Engineering}, 2021.

\bibitem{zhang2019pattern}
Zhenyu Zhang, Zhen Cui, Chunyan Xu, Yan Yan, Nicu Sebe, and Jian Yang.
\newblock Pattern-affinitive propagation across depth, surface normal and
  semantic segmentation.
\newblock In {\em Proceedings of the IEEE/CVF Conference on Computer Vision and
  Pattern Recognition}, pages 4106--4115, 2019.

\bibitem{zhao2017pyramid}
Hengshuang Zhao, Jianping Shi, Xiaojuan Qi, Xiaogang Wang, and Jiaya Jia.
\newblock Pyramid scene parsing network.
\newblock In {\em Proceedings of the IEEE Conference on Computer Vision and
  Pattern Recognition}, pages 2881--2890, 2017.

\bibitem{zhou2015learning}
Bolei Zhou, Aditya Khosla, Agata Lapedriza, Aude Oliva, and Antonio Torralba.
\newblock Learning deep features for discriminative localization.
\newblock In {\em Proceedings of the IEEE Conference on Computer Vision and
  Pattern Recognition}, pages 2921--2929, 2016.

\bibitem{zhou2017unsupervised}
Tinghui Zhou, Matthew Brown, Noah Snavely, and David~G Lowe.
\newblock Unsupervised learning of depth and ego-motion from video.
\newblock In {\em Proceedings of the IEEE Conference on Computer Vision and
  Pattern Recognition}, pages 1851--1858, 2017.

\bibitem{zou2020learning}
Yuliang Zou, Pan Ji, Quoc-Huy Tran, Jia-Bin Huang, and Manmohan Chandraker.
\newblock Learning monocular visual odometry via self-supervised long-term
  modeling.
\newblock In {\em European Conference on Computer Vision}, pages 710--727,
  2020.

\end{thebibliography}
}

\end{document}


\title{Supplementary Material}
\author{Nitin Bansal, Pan Ji, Junsong Yuan, Yi Xu\\
}
\maketitle
\paragraph{AffineMix Data Augmentation}
In this subsection we illustrate the steps followed to generate augmented images using the proposed AffineMix method. We call AffineMix, an intra data augmentation technique, as we do not use any other image to create the augmented image, but use the same target frame $I$ and then apply random affine transformation on a selected movable object to get a new image $I'$ as shown in algorithm ~\ref{alg:Data_aug_intra}.

\begin{algorithm}
\caption{Intra Data Augmentation}
\label{alg:Data_aug_intra}
\begin{algorithmic}[1]
\State \textbf{\# Intra Data Augmentation step:}
\State $Scale \gets random (0.5, 1.5)$
\State $t_x = (1.0 - 1/scale)) * (o_x)$ \Comment{Offset along x}
\State $t_y = (1.0 - 1/scale)) * (o_y)$ \Comment{Offset along y}
\State \textbf{ }
\State \textbf{\# Affine transform according to Scale:}
\State $I_a = aff\_transform(I, 1/scale, [t_x, t_y])$
\State $L_a = aff\_transform(L, 1/scale, [t_x, t_y])$
\State $S_a = aff\_transform(S, 1/scale, [t_x, t_y])$
\State $D_a = aff\_transform(D, 1/scale, [t_x, t_y])$
\State $D_a = Scale * D\_a$
\State \textbf{ }
\State \textbf{\# Generate new image}
\State $M = D_a \leq D$ \Comment{foreground mask}
\State $I' = M \odot I_a + (1-M) \odot I$ \Comment{Augmented Image}
\State $L' = M \odot L_a + (1-M) \odot I$ \Comment{Augmented Label}
\State $S' = M \odot L_a + (1-M) \odot S$ \Comment{Augmented Softmax}
\end{algorithmic}
\end{algorithm}

\begin{figure}[ht]
	\begin{center}
		\includegraphics[scale=0.59]{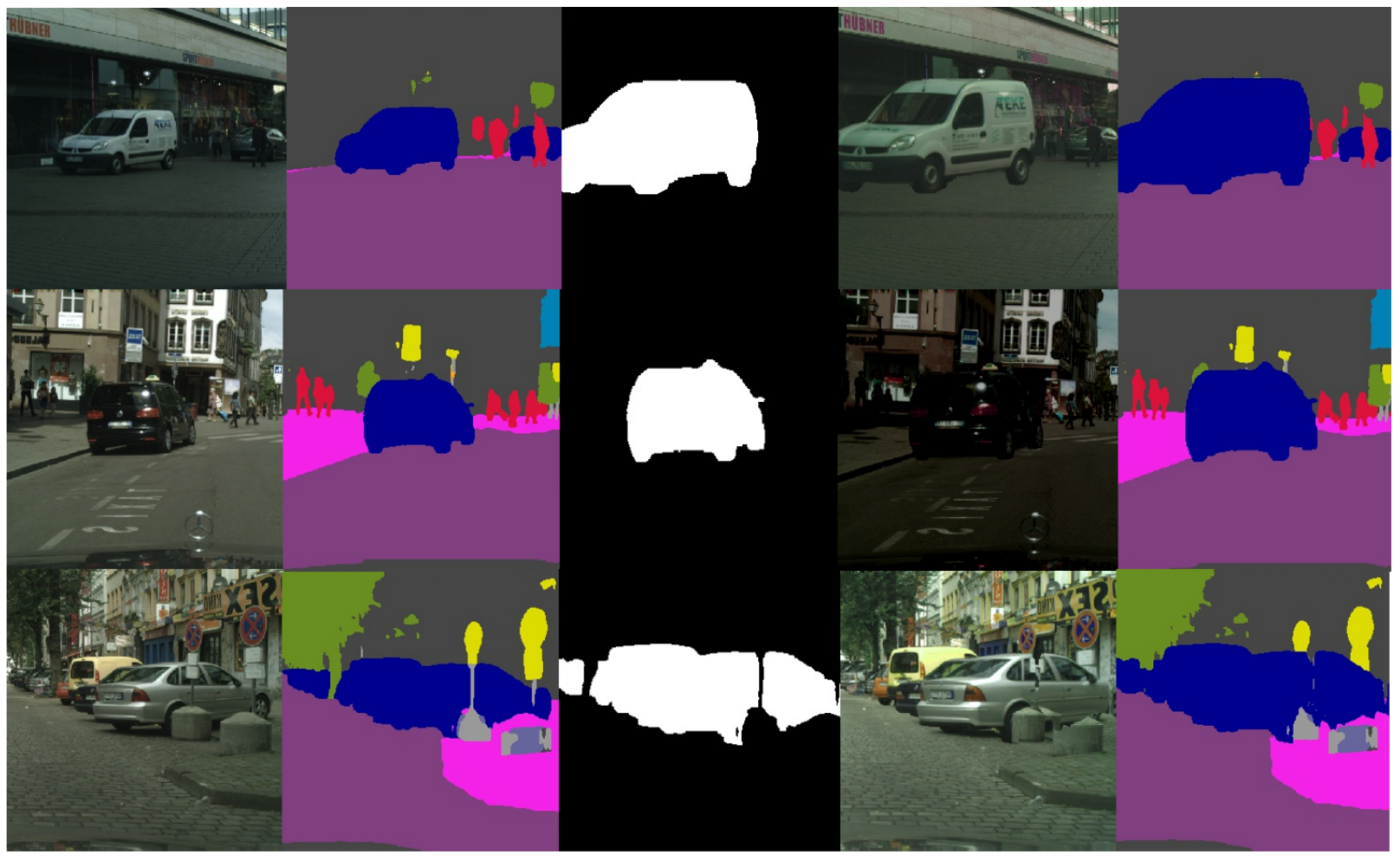}
	\end{center}
    \vspace{-1em}
	\caption{\textbf{Example Images AffineMix}:Left-to-right: Original Image, GT Labels, Scaled foreground mask, Augmented Image, New label space}
	\label{fig:intra_aug_fexp}
        \vspace{-1em}
\end{figure}

We choose a random scale for depth which lies in range(0.5, 1.5), and then perform affine transform on the selected moving object mask $M$ to achieve the new augmented frames, which are then used during semi-supervised semantic segmentation learning. Fig.~\ref{fig:intra_aug_fexp} provides further qualitative examples of proposed data augmentation scheme.

\paragraph{Orthogonality}
In our ablation study, we confirm that enforcing orthogonality has a positive impact on both depth and semantics performance.
\begin{figure*}[t]
	\begin{center}
		\includegraphics[scale=0.16]{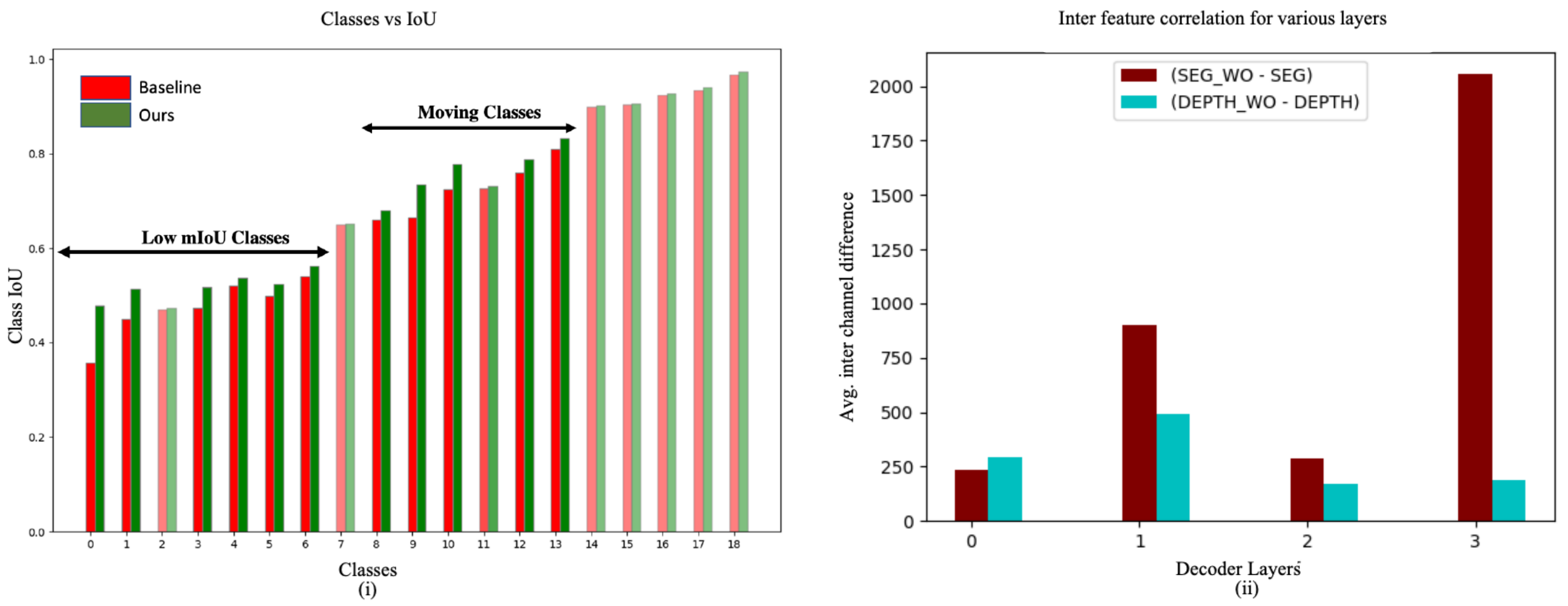}
	\end{center}
   \vspace{-1.5em}
	\caption{(i) \textbf{Class Specific mIoU:} Comparative improvement in IoU numbers in increasing order. (ii) Figure shows the difference between the average correlation between all features for semantics and depth decoder, for without (WO) OR and with OR.}
	\label{fig:mIoU_comp}
	\vspace{-0.5em}
\end{figure*}
Here we show its effect on features of different layers of the depth and semantics decoder module, once the training has ended. Suppose a layer of depth decoder has a dimensionality of (BxCxHxW), where B,C,H and W represents batch size, number of channels, height and width respectively. We estimate an average inter-channel correlation, according to following algorithm:
\begin{algorithm}
\caption{Inter Channel Correlation}
\label{alg:inter_corr}
\begin{algorithmic}[1]
\Require $layer$ \Comment{Intermediate CNN feature layer}
\State $B,C,H,W =  layer.dim$
\While{$c \leq C$}
\State $norm =  0.0$
\While{$i \leq C$}
\If{$i$ = $c$}
\State $continue$
\Else
\State $corr = layer[:,i,:,:] * layer[:,c,:,:]^T$
\State $norm += l1\_norm(corr)$
\EndIf
\EndWhile
\State $norm = norm/C$ \Comment{average norm per layer}
\EndWhile
\end{algorithmic}
\end{algorithm}

We calculate the average norm as specified in Algo.~\ref{alg:inter_corr}, for all decoder layers for both the tasks, with and without OR(orthogonal regularization). In Fig.~\ref{fig:mIoU_comp} (ii), we plot average difference of correlation for four different layers of semantics and depth decoder module, without and with OR. As seen in the plot, we observe a positive difference all along for both semantics and depth module, suggesting greater inter-channel independence.

\begin{figure*}[!hbt]
    \begin{center}
		\includegraphics[scale=0.3]{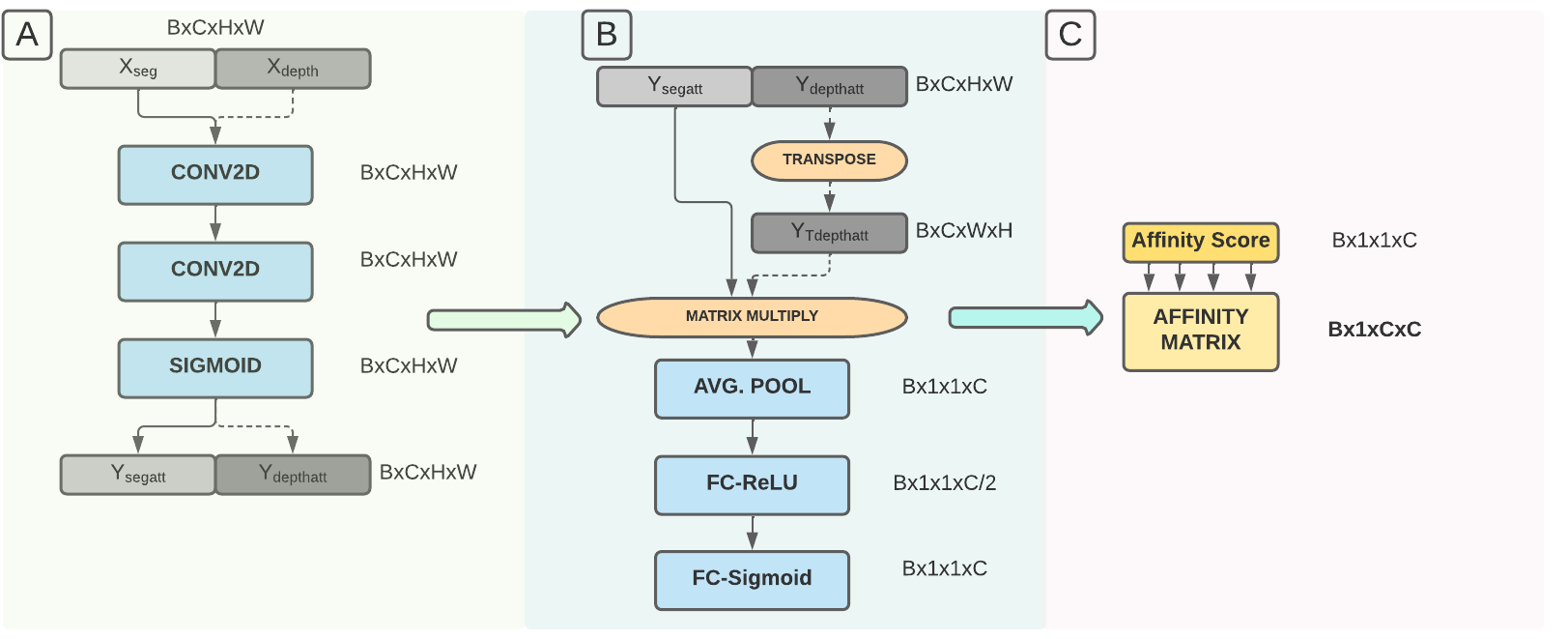}
	\end{center}
	\caption{Building blocks of the \textbf{Cross Channel Affinity Module}.}
	\label{fig:ccam_module}
	\vspace{1em}
\end{figure*}

\paragraph{CCAM Module}
As seen in Fig.~\ref{fig:ccam_module}, architecture of CCAM module can be divided in to three blocks namely \textbf{A}, \textbf{B} and \textbf{C}, which serves three different purpose. Block A acts as a spatial attention layer, where as Block B estimates cross feature correlation and follows it up with channel attention layer to give us an \textit{affinity score} per feature channel, which is then followed by Block C, which does a simple channel-wise accumulation of \textit{affinity score}  to give final Affinity Matrix. Fig.~\ref{fig:ccam_module} shows individual learnable layers part of all the three blocks in complete detail. We also present the steps followed for calculating affinity matrix and getting resultant depth and semantic features in algorithm~\ref{alg:cam}.\\
\begin{algorithm}
\caption{Cross Channel Affinity Block}\label{alg:cam}
\begin{algorithmic}[1]
\Require $X_{seg}$, $X_{depth}$ \Comment{Semantics and Depth features} 
\State \textbf{\# Estimate cross channel affinity:}
\State $Y_{segatt}     = Spatial\_Attn(X_{seg})$
\State $Y_{depthatt}   = Spatial\_Attn(X_{depth})$
\State $Y_{Tdepthattn} = Transpose(Y_{depthatt})$
\State $i \gets 0$

\While{$i < C$} \Comment{C represents number of channels}
\State $\alpha_i = Channel\_Affinity(Y_{segatt} * Y_{Tdepthattn})$
\State $C_T = C_T +  \alpha_i$ \Comment{concat across dimension}
\EndWhile
\State \textbf{ }
\State \textbf{\# Mutual features sharing between tasks:}
\State $i \gets 0$
\While{$i < C$}
\State $X_{seg}  = X_{seg} + X_{depth_i} * C_{Ti}$ \Comment{along row dimension}
\EndWhile
\State $j \gets 0$
\While{$j < C$}
\State $X_{depth} = X_{depth} + X_{seg_j} * C_{Tj}$ \Comment{along column dimension}
\EndWhile
\end{algorithmic}
\end{algorithm}

\begin{figure*}[!hbt]
	\begin{center}
		\includegraphics[scale=0.78]{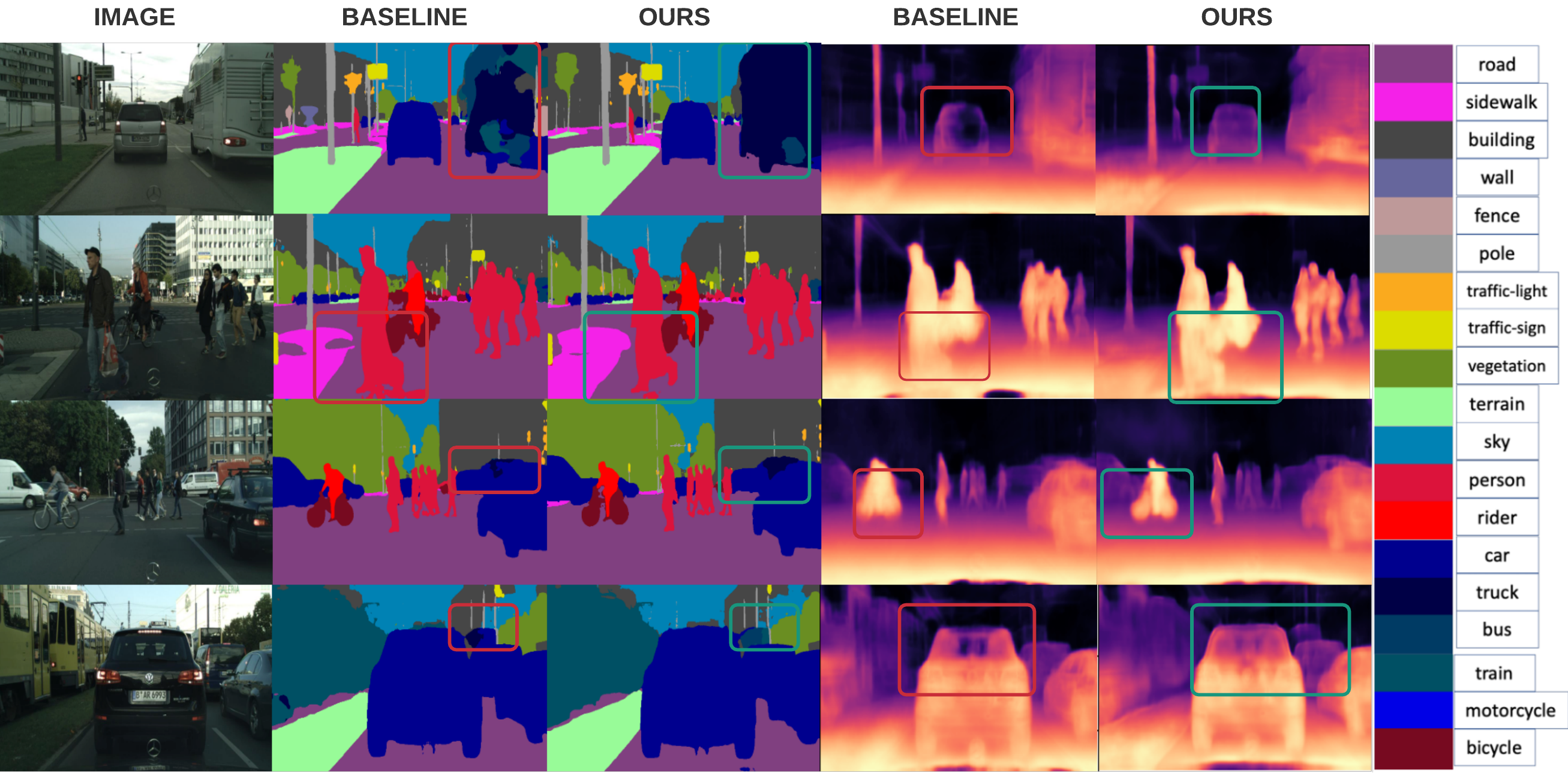}
	\end{center}
    \vspace{-1em}
	\caption{\textbf{Example Outputs:} Semantics and depth results for the same input images, showing comparative results with the baseline.}
	\label{fig:imp_seg_depth}
\end{figure*}

 \begin{figure*}[t]
    \begin{center}
        \includegraphics[scale=0.65]{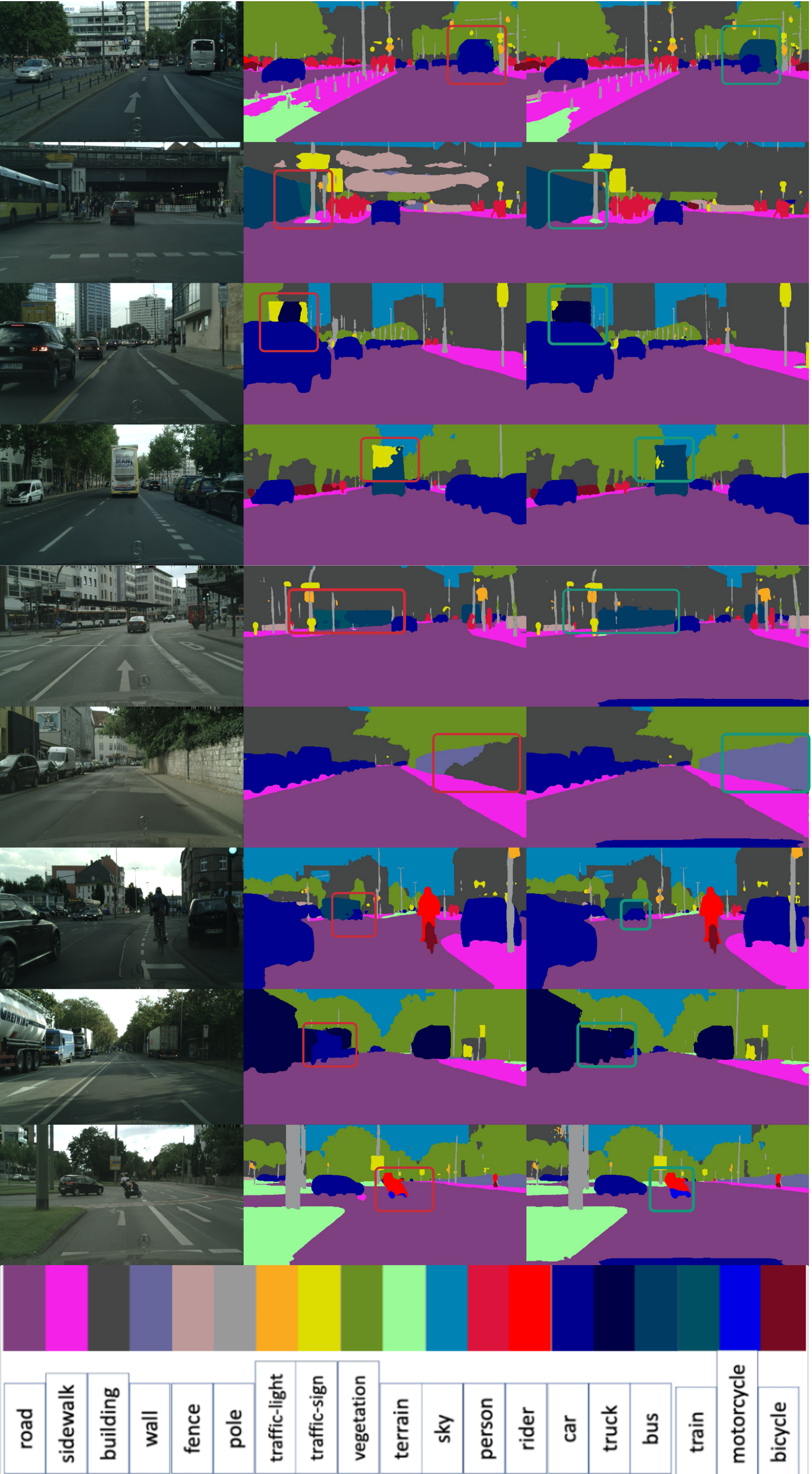}
    \end{center}
    \vspace{-1em}
    \caption{\textbf{Examples of Improved classes:} From up-to-down:(Bus-Car), (Bus-Train), (Truck-Sign), (Truck-Sign), (Bus-Train), (Wall-Building), (Bus-Car),(Truck,Car), (Rider, Motorcycle)}
        \vspace{-1em}
    \label{fig:imp_class}
\end{figure*}

\begin{figure*}[!ht]
	\begin{center}
		\includegraphics[scale=0.78]{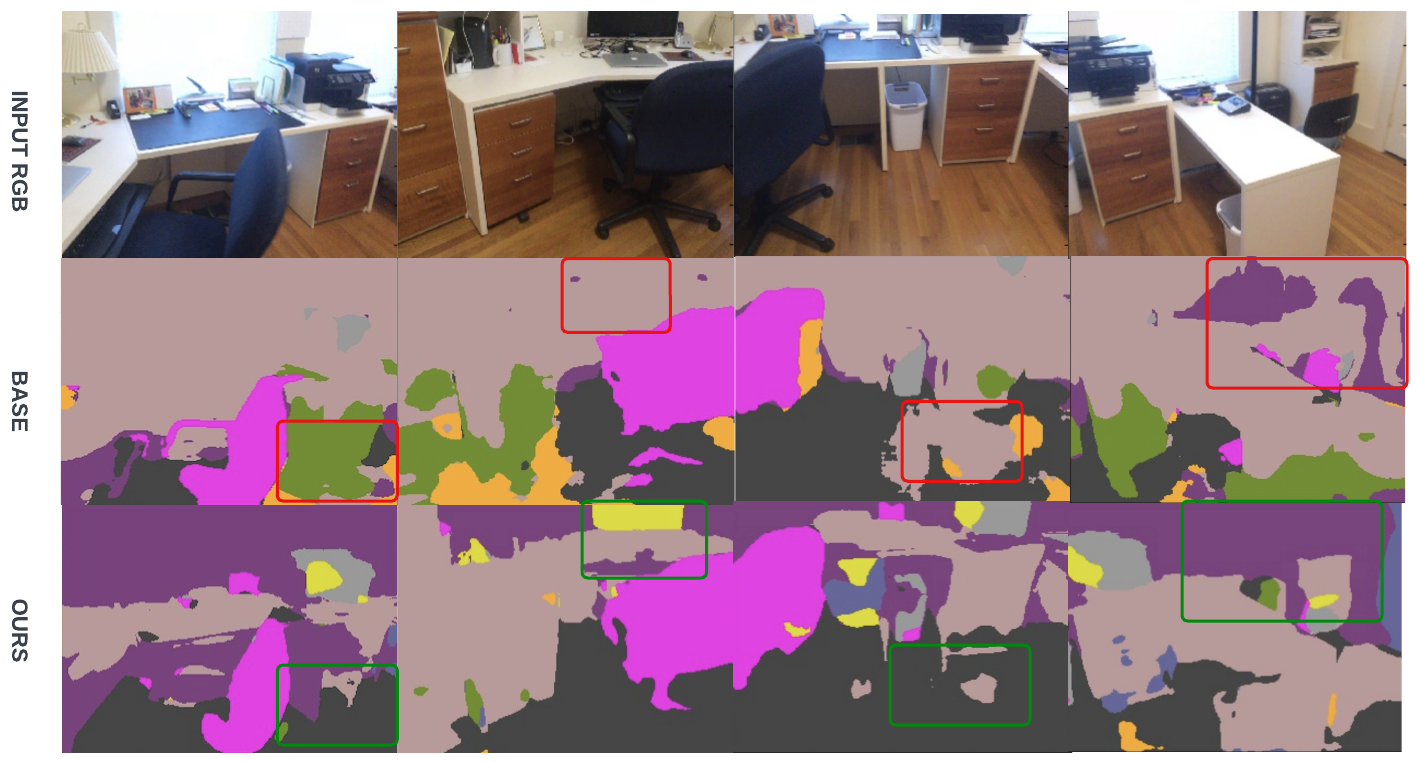}
	\end{center}
    \vspace{-1em}
	\caption{Example improvement between  Base and Our model for scene0166\_00 in ScanNet dataset. Above image shows improvement for 'Floor', 'Screen', 'Floor' and 'Wall' class respectively.}
	\label{fig:scannet_imgs}
\end{figure*}

\paragraph{Further Qualitative Results}
\label{Further_Results}
We have discussed in the main paper there are certain classes within the 19 categories present in the cityscapes dataset which have shown more improvement compared to other \textit{saturated classes}. We have marked them as \textit{low-mIoU-classes} and \textit{movable-classes} which were mainly (``rider'', ``motorcycle'', ``wall'', ``bus'', ``truck'', and ``train'') respectively. For this we plot mIoU numbers in an increasing order \footnote{Order: motorcycle, wall, fence, rider, pole, traffic-light, terrain, traffic-sign, bicycle, train, truck, person, sidewalk, bus, building, vegetation, car, sky, road.} As seen in Fig.~\ref{fig:mIoU_comp}(i), we see that most of the improvement is seen for classes belonging to \textit{low-mIoU-classes} and \textit{movable-classes}. We particularly highlight the some of the positive examples of these classes, as seen across different cities for test set of cityscapes dataset in Fig.~\ref{fig:imp_class}. We find there are quite a few examples, where the mistakes are mainly due to a mix-up in the predictions mainly concerning the labels belonging to class set of bus, car, truck and train due to obvious visual similarity. In Fig.~\ref{fig:imp_seg_depth}, we show qualitative depth and semantics results for the same set of input images, highlighting improvement obtained on both tasks. Here, it is imperative to state that it's not necessary that we see an improvement for both the tasks, over all input test images. There are few classes which show minimal improvement and also certain degenerate example images where we fail to effectively handle \textit{unknown} classes, examples of which are discussed in Sec.~\ref{limitations}. Example Qualitative result for ScanNet data can be seen in Fig. \ref{fig:scannet_imgs}\\
\paragraph{\bf Training Details:}
Our Proposed model achieves better performance with approximately the same number of parameters ($>$2000 more parameters than baseline). There is a slight increase in the number of flops as seen in Tab.~\ref{tab:train_detail}, which can be mainly attributed to the matrix multiplication step, used during the correlation calculation step in Block \textit{B} (See Fig.~\ref{fig:ccam_module}).
\begin{table*}[!ht]
    \center
     \vspace{-1em}
    \scalebox{1.0}{
	\begin{tabular}{p{1.6cm} p{1.9cm}p{1.9cm}}
		\hline
		Model& Flops$^+$ ($10^9$) & Params ($10^6$) \\
		\hline
		Baseline &  - & $87.226$ \\
		Ours & $0.27$ & $87.228$ \\
		\hline
		\vspace{-1em}
	\end{tabular} }
    \vspace{-1em}
    \caption{Comparative Training details. $+$ denotes additional numbers of flops used.}
    \vspace{-1em}
    \label{tab:train_detail}
\end{table*}
\vspace{1em}

\begin{figure*}[!ht]
	\begin{center}
		\includegraphics[scale=0.138]{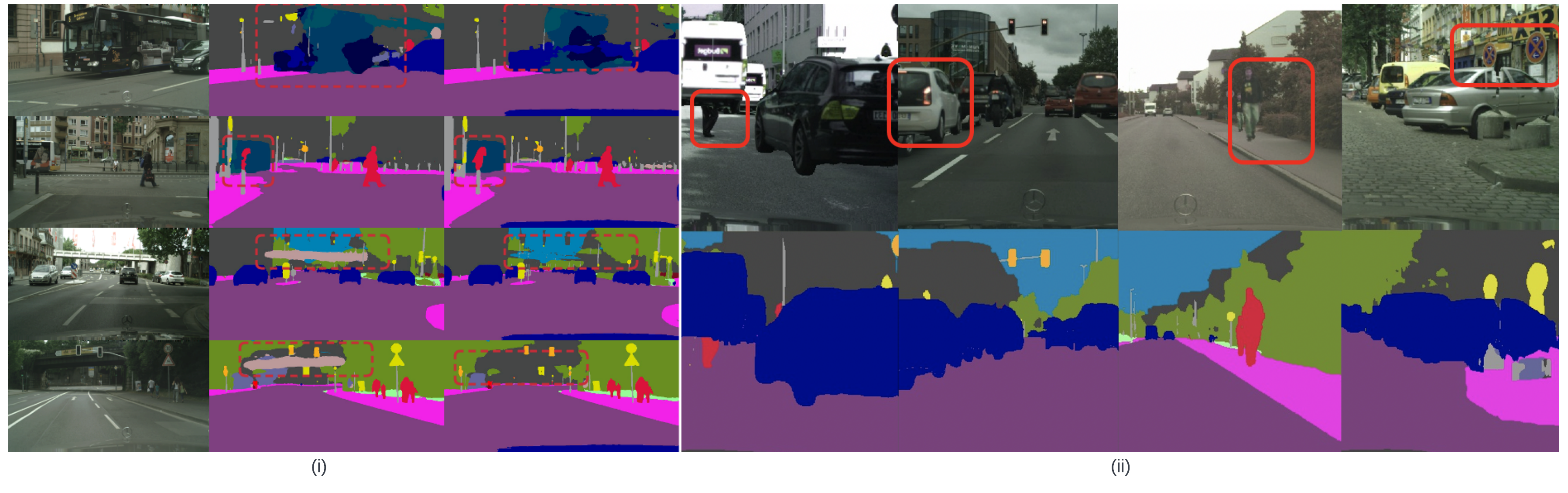}
	\end{center}
    \vspace{-1em}
	\caption{\textbf{Bad Examples:}(i) Few examples where the model gets confused and makes the wrong prediction. Left-to-Right: Input image, Baseline, Ours.}
	\label{fig:bad_exam}
\end{figure*}

\paragraph{More Ablation Experiments}
\label{Ablation}
\begin{table*}[!t]
\scalebox{1.0}{
\setlength{\tabcolsep}{12.4pt}{
\begin{tabular}{l|c|cccccc}
\hline
Model       &     \multicolumn{1}{c|}{Seg. Metrics} & \multicolumn{6}{c}{Depth Metrics}\\
\hline
& mIoU$\uparrow$ & AbsRel$\downarrow$ &SqRel$\downarrow$& RMSE$\downarrow$& a1$\uparrow$ & a2$\uparrow$ & a3$\uparrow$ \\
\hline 
Ours (CCAM) &{69.35} &{0.142}&{1.653}&{7.230}&\textbf{0.824}&\textbf{0.957}&{0.988}\\
\hline
Ours + OR  &{69.73} &{0.143}&{1.612}&{7.433}&{0.817}&{0.954}&{0.987}\\
\hline
Ours + CA &{69.13} &\textbf{0.140}&\textbf{1.638}&{7.317}&{0.819}&{0.954}&\textbf{0.988}\\
\hline
Ours + CA + AM &{69.74} &{0.144}&{1.471}&\textbf{7.158}&{0.812}&{0.955}&{0.988}\\
\hline
Ours + CA + OR &\textbf{69.77} &{0.148}&{1.666}&{7.309}&{0.789}&{0.953}&{0.987}\\
\hline
\end{tabular}}}
\vspace{1mm}
\caption{Ablation experiments showing comparative mIoU and depth results for Cityscape dataset. CA: Color Aug, OR: Orthogonal Regularization, AM: Affine Mix}
\label{tab:compare_methods_cityscapes}
\vspace{-1mm}
\end{table*}
\begin{table*}[!t]
    \center
    \scalebox{0.9}{
	\begin{tabular}{p{1.6cm} p{1.5cm} p{1.5cm}p{1.5cm}p{1.5cm} p{1.5cm}p{1.5cm}p{1.5cm}p{1.5cm} p{1.5cm}}
		\hline
		Model & Wall & Chair & Floor & Door & Table & Box & Screen & Cabinet & mIoU \\
		\hline
		{Base} & {71.04} & \textbf{62.90} & \textbf{64.50} & {10.60} & {46.50}  & {8.70} & {22.00} & {26.20} & {39.50}\\
		\hline
		{Ours(CCAM)} & \textbf{72.40} & {60.10} & {62.00} & \textbf{21.30} & \textbf{48.30}  & \textbf{11.60} & \textbf{25.10} & \textbf{31.80} & \textbf{41.57}\\
		\hline
	\end{tabular} }
	\vspace{1.0mm}
    \caption{Table presents the comparative performance of 8 different classes in ScanNet dataset between the baseline and CCAM enabled model.}
    \label{tab:scannet_class_ious}
    \vspace{-1mm}
\end{table*}
\renewcommand{\arraystretch}{1.2} 
\begin{table*}[!t]
    \center
    \scalebox{0.8}{
	\begin{tabular}{p{0.73cm} p{0.6cm} p{0.6cm}p{0.6cm}p{0.6cm} p{0.6cm}p{0.6cm}p{0.6cm}p{0.6cm} p{0.6cm}p{0.6cm} p{0.6cm} p{0.6cm}p{0.6cm}p{0.6cm} p{0.6cm}p{0.6cm}p{0.6cm}p{0.6cm}p{0.6cm}p{0.6cm}p{0.6cm}}
		\hline
		Model & 1 & 2 & 3 & 4 & 5 & 6 & 7 & 8 & 9 & 10 & 11 & 12 & 13 & 14 & 15 & 16 & 17 & 18 & 19 & mIou  \\
		\hline
		{Base} & {35.74} & {44.92} & {46.97} & {47.21} & {52.11}  & {49.88} & {54.12} & {65.02} & {66.07} & {66.60} &{72.56}&{72.66}&{76.00}&{80.93}&{89.93}&{90.41}&{92.42}&{93.34}&{96.79} &{68.09}\\
		\hline
		{Ours} & {40.53} & {52.60} & {47.97} & {53.11} & {53.63}  & {53.30} & {56.14} & {66.38} & {67.47} & {75.17} &{77.08}&{73.31}&{78.14}&{83.91}&{90.35}&{90.57}&{93.03}& {93.78}&{97.12}&{70.72}\\
		\hline
		{Change} & {4.79$\uparrow$} & {7.68$\uparrow$} & {1.00$\uparrow$} & {5.91$\uparrow$} & {1.52$\uparrow$}  & {3.42$\uparrow$} & {2.02$\uparrow$} & {1.36$\uparrow$} & {1.40$\uparrow$} & {8.57$\uparrow$} &{4.52$\uparrow$}&{0.65$\uparrow$}&{2.14$\uparrow$}&{2.98$\uparrow$}&{0.42$\uparrow$}&{0.16$\uparrow$}&{0.61$\uparrow$}& {0.44$\uparrow$}&{0.33$\uparrow$}&{2.63$\uparrow$}\\
		\hline
	\end{tabular} }
	\vspace{1.5mm}
    \caption{Table presents the comparative performance for all 19 classes in Cityscapes dataset. Order of class is: motorcycle, wall, fence, rider, pole, traffic-light, terrain, traffic-sign, bicycle, train, truck, person, sidewalk, bus, building, vegetation, car, sky, and road.}
    \label{tab:cityscapes_class_ious}
    \vspace{-3mm}
\end{table*}
\renewcommand{\arraystretch}{1.0}
We ran further set of ablation experiments to verify the efficacy and impact of each individual component. Table ~\ref{tab:compare_methods_cityscapes} presents the result for all conducted experiments in detail. Observations are consistent with that seen in the main paper. Inclusion of CA helps in further consolidating the gain for depth estimation achieving the best absolute relative error of \textbf{0.140}, whereas adding OR and AM module helps in improving the base result by \textbf{0.38} and \textbf{0.39} mIoU numbers respectively. We do see a slight dip in mIoU while using CA, but when combined with AM/OR it always performs better even for semantic segmentation.
\paragraph{Quantitative Results} We present class wise mIoU number for both ScanNet and Cityscapes dataset as shown in Table~\ref{tab:scannet_class_ious} and ~\ref{tab:cityscapes_class_ious}. For ScanNet, we see improvement for all the classes except for class \textit{Chair} and \textit{Floor}, Where as for Cityscapes, we see improvement across all the classes, albeit to varying degree, with \textit{low-mIoU-classes} and \textit{movable-classes} seeing the majority of gain.

 \begin{figure*}[t]
    \begin{center}
        \includegraphics[scale=0.60]{Imgs/Class Hierarchy.png}
    \end{center}
    \vspace{-1em}
    \caption{Above figure shows hierarchical class structure, with 8 parent classes. This also shows the mapping between classes defined in NYU~\cite{silberman2012indoor} and ScanNet dataset.}
        \vspace{-1em}
    \label{fig:class_hierar}
\end{figure*}

\paragraph {Dataset:}
\textbf{Cityscapes}
Cityscapes~\cite{cordts2016cityscapes} consist of 2,975 training and 500 validation images with ground-truth semantic segmentation labels, collected from 21 different European cities. For semi-supervised segmentation, we use only 2,975 labeled training images, which are randomly split into a labeled and an unlabeled subset in accordance with number of labelled images being used for training.

\noindent \textbf{ScanNet} During our experiment with ScanNet~\cite{dai2017scannet}, we mainly focus on scenes which are from \textit{Living room}, \textit{Bedroom} and \textit{Office}, which mainly consist of 8 (parent) classes as shown in fig.~\ref{fig:class_hierar}, to make joint-training bit smoother. Using which we get a new train/val/split of 76/15/11 scenes respectively. All other classes were set to \textit{ignore class} during both training and evaluation. Also, since we have the ground-truth pose for ScanNet, we use it during depth prediction. We follow similar data preprocessing steps for ScanNet during training and evaluation as done for Cityscapes. 
\paragraph{Model Architecture}
It comprises of a shared encoder network which is ResNet101~\cite{he2015deep} with output stride as 16. We use two different decoder modules for semantics and depth respectively, which is a combination of Deeplabv3~\cite{chen2017deeplab} with a U-Net~\cite{ronneberger2015unet} decoder.
ASPP Blocks~\cite{he2015deep} with dilation rates of 6, 12, and 18 are used for aggregating encoder features of different scales. U-Net~\cite{ronneberger2015unet} decoder has five upsampling blocks with skip connections, with output channel as 256, 256, 128, 128, and 64 respectively.

\section{Limitations}
\label{limitations}
In this section, we briefly discuss specific limitations of our method. We start with illustrating examples of input images, where we fail to handle \textit{unseen} and \textit{confusing} scenarios and make wrong predictions.
We find there are specific cases, as shown in Fig.~\ref{fig:bad_exam}(i) (row 1, 2), where picture/painting and transparent glass on the \textit{bus}'s exterior makes it difficult for the model to classify \textit{bus} correctly. We also find that images containing flyover or overhead bridges (See Fig.~\ref{fig:bad_exam}(i) (row 3, 4))  are either misclassified or missed. During our experiment, we also find that there is little impact mIoU-wise on classes such as traffic-sign, fence, and vegetation. In Figure~\ref{fig:bad_exam} (ii), we highlight certain examples of augmented images that need not necessarily obey the geometry of the scene and have hanging, perspective and truncated artifacts (due to occluding stationary object) respectively. These are particularly seen in cases that involve large depth changes during the data augmentation and is enhanced due to using an average camera intrinsics for all scenes. Another direction of improvement would be to further study effects arising from incorrect lighting arising due to proposed data augmentation, which needs to be investigated to understand its impact on both tasks. We look forward to work towards finding meaningful answers for the aforementioned scenarios in future work.
{\small
\bibliographystyle{ieee_fullname}
\bibliography{egbib}
}